\documentclass[manuscript,screen]{acmart}
\usepackage[utf8]{inputenc} 
\usepackage[T1]{fontenc}    
\usepackage{hyperref}       
\usepackage{url}            
\usepackage{booktabs}       
\usepackage{amsfonts}       
\usepackage{nicefrac}       
\usepackage{microtype}      
\usepackage{xcolor}         
\usepackage{float}
\usepackage{amsmath}

\usepackage{graphicx}  
\usepackage{bm}
\usepackage{algorithm, algorithmic}
\usepackage{caption}
\usepackage{subfigure}
\usepackage{makecell}
\usepackage{wrapfig}
\usepackage{stfloats}
\usepackage{bbm}
\usepackage{multirow}
\usepackage{makecell}
\usepackage{colortbl}

\AtBeginDocument{%
  \providecommand\BibTeX{{%
    \normalfont B\kern-0.5em{\scshape i\kern-0.25em b}\kern-0.8em\TeX}}}

\setcopyright{acmcopyright}
\copyrightyear{2018}
\acmYear{2018}
\acmDOI{XXXXXXX.XXXXXXX}

\acmConference[Conference acronym 'XX]{Make sure to enter the correct
  conference title from your rights confirmation emai}{June 03--05,
  2018}{Woodstock, NY}
\acmPrice{15.00}
\acmISBN{978-1-4503-XXXX-X/18/06}

\begin{document}

\title{Building Shortcuts between Distant Nodes with Biaffine Mapping for Graph Convolutional Networks}

\author{Acong Zhang}
\authornote{Both authors contributed equally to this research.}
\email{zac1328682511@gmail.com}
\affiliation{%
  \institution{School of Computer Science, Southwest Petroleum University}
  \streetaddress{P.O. Box 610500}
  \city{Chengdu}
  \state{Sichuan}
  \country{China}
}
\author{Jincheng Huang}
\authornotemark[1]
\email{huangjc0429@gmail.com}
\affiliation{%
  \institution{School of Computer Science, Southwest Petroleum University}
  \streetaddress{P.O. Box 610500}
  \city{Chengdu}
  \state{Sichuan}
  \country{China}
}

\author{Ping Li}
\authornote{Corresponding author.}
\authornotemark[2]
\orcid{0000-0002-8391-6510}
\affiliation{%
  \institution{School of Computer Science, Southwest Petroleum University}
  \streetaddress{P.O. Box 610500}
  \city{Chengdu}
  \state{Sichuan}
  \country{China}}
\email{dping.li@gmail.com}

\author{Kai Zhang}
\affiliation{%
  \institution{School of Computer Science and Technology, East China Normal University}
  \streetaddress{Zhongshan North Road}
  \city{Shanghai}
  \country{China}}
\email{kzhang@cs.ecnu.edu.cn}

\renewcommand{\shortauthors}{XXXX.}

\begin{abstract}
  Multiple recent studies show a paradox in graph convolutional networks (GCNs), that is, shallow architectures limit the capability of learning information from high-order neighbors, while deep architectures suffer from over-smoothing or over-squashing. To enjoy the simplicity of shallow architectures and overcome their limits of neighborhood extension, in this work, we introduce \emph{Biaffine} technique to improve the expressiveness of graph convolutional networks with a shallow architecture. The core design of our method is to learn direct dependency on long-distance neighbors for nodes, with which only one-hop message passing is capable of capturing rich information for node representation. Besides, we propose a multi-view contrastive learning method to exploit the representations learned from long-distance dependencies. Extensive experiments on nine graph benchmark datasets suggest that the shallow biaffine graph convolutional networks (BAGCN) significantly outperforms state-of-the-art GCNs (with deep or shallow architectures) on semi-supervised node classification. We further verify the effectiveness of biaffine design in node representation learning and the performance consistency on different sizes of training data.
\end{abstract}




\maketitle

\section{INTRODUCTION}
Recent years have witnessed the rapid progress of graph neural networks (GNNs)~\cite{tkde_graphservey,tkde_graphservey_2}, a class of neural networks that can learn from graph-structured data. Since the early success of GCN models~\cite{cheyGCN2016,kipf2017gcn} on node classification, many variants of GCNs have been widely applied in social network analysis~\cite{socialnetwork}, natural language processing~\cite{biaffine-1}, recommendation systems~\cite{lightgcn,pinsage} and semi-supervised learning~\cite{deeper,semi-graph}.  In semi-supervised learning setting where a very small number of labeled data is used to predict the classes of unlabeled data, GNNs are endowed with the ability to capture the correlation between unlabeled and labeled data by message passing on the graph.

\begin{figure*}[htpb]
	\centering
	\includegraphics[scale=0.35]{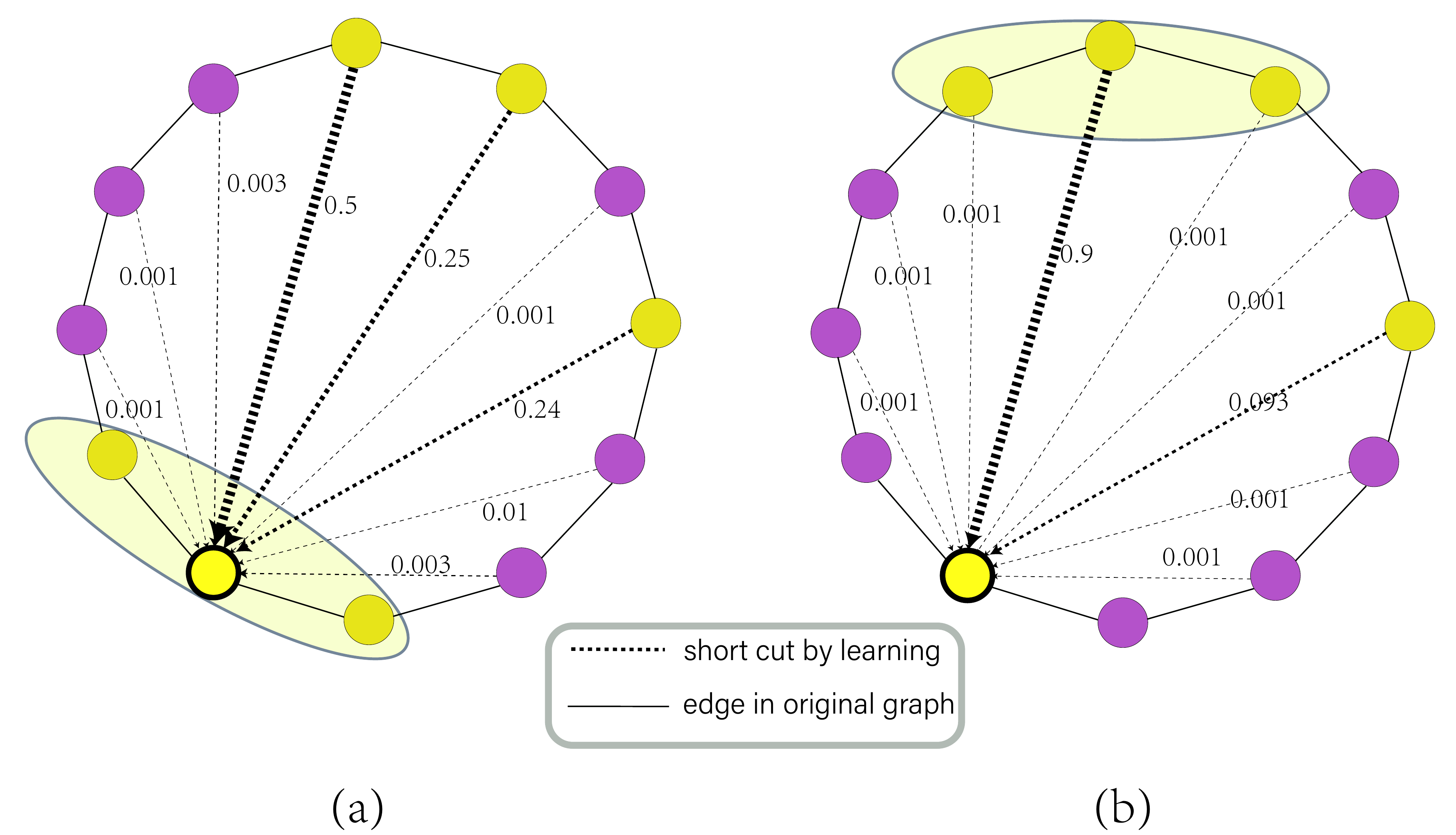}
	\caption{A toy example to illustrate the biaffine effects. The shortcuts are denoted by dashed lines and the thickness of the dashed lines indicates the dependency degree. The nodes labeled in similar color are assumed to have the same class label. The node in question is highlighted in bright yellow. (a) long range dependency at local field level. 
		(b) long range dependency at ego level.
	}
	\label{fig:biaffine_gnn}
\end{figure*}

However, ``every coin has two sides". Message passing based GCNs is the simplification of Chebyshev truncation~\cite{mixhop2019}, which means that in each convolution layer, GCNs can only leverage the immediate neighboring nodes, missing the information of higher-order neighbors. Though stacking multiple convolution layers can be used to capture long distance relationships, it is observed in prior study~\cite{oversmooth} that the features of all nodes will become indistinguishable with the increase of convolution layers, which can have significant impacts on performance. This is the well-known \emph{over-smoothing} phenomenon emerging in deep GCNs. To benefit from deep architecture, some techniques borrowed from deep learning (e.g., residual connection~\cite{gcnII, jknet} and layer normalization~\cite{pairnorm}) are used to alleviate the over-smoothing problem. Even so, deep architecture of GNNs has another side effect revealed by the recent studies~\cite{oversquashing,oversquashing22}. In fact, as the number of layers increases, the number of nodes in each node's receptive field grows exponentially, resulting in \emph{over-squashing}: information from the exponentially-growing receptive field is compressed into fixed-length node vectors~\cite{oversquashing}. Since the over-squashing is considered to be related to the bottleneck of graph structure where the scale of neighborhood exponentially increase with deep convolution, graph-rewiring has been adopted to mitigate the bottleneck and the optimal GNN models on the rewired graph are generally one-layer architecture~\cite{oversquashing22}.

The above observations suggest that \emph{both phenomena are derived from deep layers}, which raises the question: whether it is possible to adaptively learn the feature dependence of a node on both local and global structure of the graph without resort to deep architectures. To achieve this, we explore a shallow architecture for graph convolutional networks that attentively build shortcuts (i.e., long-range dependency) between nodes and their differing long-range neighbors. This architecture allow the messages passing through shortcuts so that higher-order neighbors will directly contribute to the representation learning of the nodes in question, which can avoid long-term message passing in deep graph networks. The shortcuts, or equivalently, long-range dependency, are learned at two levels, namely, local field level and ego level, by leveraging biaffine technique, which is demonstrated with a toy network ( shown in Figure~\ref{fig:biaffine_gnn}). Specifically, in Figure~\ref{fig:biaffine_gnn}(a), the feature of 1-hop neighborhood local field of an ego node is cast as query and the features of all the nodes are cast as keys, so the shortcuts (denoted by dashed lines) are dynamically determined by the similarity between "query" and "keys".
Similarly, Figure~\ref{fig:biaffine_gnn}(b) shows how another type of shortcuts can be learned by viewing each ego node as query and the first-order neighborhood of ego nodes. The intuition behind this design is that, as the local field (e.g., 1-hop neighborhood) holds the clue to the local structural and featural information about ego node, if node's feature (or local field) is similar to the local field (or feature) of the other, then it is generally considered that this node is strongly dependent of the other one. It is expected to play a role when the nodes of the same class are distributed in discontiguous subgraphs (as the toy example shows). 

In order to fully exploit the long-range dependencies at two levels (i.e., local field level and ego level), we devise a multi-view contrastive learning scheme to make predictions on top of the relations corresponding two levels.
Our contributions are summarized as follows:
\begin{itemize}
	\item We characterize dynamical correlations between nodes using the information about local field of nodes and introduce biaffine mapping into graph convolutional networks to establish shortcuts between nodes at two levels for the first time. This technique makes it possible to capture long-range interactions more directly and pass messages more efficiently, thereby reducing the demand for the number of graph convolution layers.
	\item We propose a multi-view contrastive learning model for node classification on the base of two levels' dependency relations, which applies a one-layer graph convolutional network and a fully-connected network to coordinately learn node labels from two views respectively.
	\item Besides the evaluation of node classification performance, we explore the robustness of our method when faced with noise and test the expressiveness of BAGCN using the proposed framework when training data is quite scarce.
\end{itemize}

The rest of the paper is organized as follows. In section 2, we introduce notations and the related work. In section 3, we demonstrate the proposed framework of biaffine graph convolutional network. Section 4 reports experimental results and the last section concludes the paper.
\section{Preliminary and Related work}
\label{sec:prelimit}
Let \bm{$G = (V,E)$} be an undirected and unweighted graph with nodes in $V$ and edges in $E$, the nodes are described by the feature matrix~ $\boldsymbol{X} \in\mathbb{R}^{n\times f}$, where $f$ denotes the number of features per node and $n$ is the number of nodes. Each node is associated with a class label, which is depicted in the label matrix $\boldsymbol{Y} \in\mathbb{R}^{n\times C}$ with a total of $C$ classes. We represent the graph by its adjacency matrix $\boldsymbol{A} \in \{0,1\}^{n\times n}$, with each element $A_{ij}=1$ indicating that there exists an edge between $v_i$ and $v_j$, otherwise $A_{ij} = 0$. In particular, the graph with self-loops is denoted by $\boldsymbol{\tilde{A}} = \boldsymbol{A} + \boldsymbol{I_{n}}$, whose associated degree matrix of \bm{$\tilde{A}$} is \bm{$\tilde{D}$}.

\noindent\textbf{Semi-Supervised Learning for Graphs.}  Denote the labeled set by \bm{$Y^{L}$} and unlabeled data by \bm{$Y^{U}$}, the goal of semi-supervised learning in graph setting is to leverage node features \bm{$X^{U}$} of the unlabeled nodes, \bm{$X^{L}$} of the labeled nodes and the available label \bm{$Y^{L}$} to infer the label \bm{$Y^{U}$} of the unlabeled nodes, which can be formulated as an optimization problem: \bm{$F(G, X^{U}, X^{L}, Y^{L})\rightarrow Y^{U}$} for some $F$ that makes the prediction as accurate as possible. Most of the existing work implements this formula through message passing mechanism~\cite{mpnn} and Laplacian regularization~\cite{laplacianL1,laplacianL2,laplacianL3,laplacianL4}. Recent advances indicate the impressive power of graph neural networks in semi supervised learning tasks with graph structure, which will be demonstrated below.

\noindent\textbf{Graph Neural Networks.} Graph neural networks(GNNs)~\cite{2008gnn, kipf2017gcn} fill in the gaps for traditional neural networks (e.g., convolutional neural networks and recurrent neural networks) in dealing with irregular data, namely, graphs. The main idea of GNNs is to utilize the structure of graph data to diffuse the information on nodes with some specific rules. For example, one of the very starting work Kipf's GCN~\cite{kipf2017gcn} performs message passing in the light of the Laplacian matrix, where the two-layer GCN is represented as $\boldsymbol{\widehat{Y}} = softmax(\boldsymbol{\widehat{A}}\sigma(\boldsymbol{\widehat{A}XW^{(0)}})\boldsymbol{W^{(1)}})$, where \bm{$\widehat{A}=\tilde{D}^{-\frac{1}{2}}\tilde{A}\tilde{D}^{-\frac{1}{2}}$} is the symmetric normalized adjacency matrix (i.e., laplacian matrix), the activation $\sigma(\cdot)$ uses ReLU function, and \bm{$W^{(0)}$} and \bm{$W^{(1)}$} are the weight matrices of linear transformation.
There are many variants of GCNs in the literature that have shown strong performance in semi-supervised node classification task, such as Kipf's GCN~\cite{kipf2017gcn}, GAT~\cite{2018gat}, APPNP~\cite{appnp}, DAGNN~\cite{dagnn}, GRAND~\cite{grand} and so on and so forth. Most of these models achieve best performance when using two or three convolution layers and show dramatic performance degradation for more layers. A more model SUGRL~\cite{SUGRL} leverages feature shuffling and graph sampling to explore the complementary information between structural information and neighborhood information to expand interclass variation  in unsupervised setting. MixHop~\cite{mixhop2019} is one successful attempt to receive the information of multi-level neighbors with shallow architecture. But it treats all higher-order neighbors equally and simply concatenate the nodes in the ego-graph. Moreover, while possible in theory, concatenation mixing encounters practical difficulties in capture various long-range neighbors. It is noteworthy that many real-world graphs are usually large. To make GNN model scaleable, neighbor sampling is proposed in GCNs, e.g., GraphSAGE~\cite{graphSAGE2017}, FastGCN~\cite{fastgcn2018} and AS-GCN~\cite{asgcn}.

\noindent\textbf{Bi-Affine Mapping.} Biaffine is a special class of projection transformations, which can preserve the collinearity (i.e., all points lying on a line initially still lie on the line after transformation) and ratios of distances (e.g., the midpoint of a line segment remains the midpoint after transformation)~\cite{affine-1,affine-2}. Affine transformation of a vector can be described as $\boldsymbol{\vec{y}} = \boldsymbol{S\vec{x}} + \boldsymbol{\vec{b}}$, where \bm{$S$} represents the linear transformation of vector $\boldsymbol{\vec{x}}$ and $\boldsymbol{\vec{b}}$ is the translation transformation term. Basically, biaffine is the affine transformations of \bm{$\vec{x}$} to \bm{$\vec{y}$} and \bm{$\vec{y}$} to \bm{$\vec{x}$} at the same time. The early work that introduce biaffine mapping into deep learning is to learn token dependency relation prediction~\cite{biaffine-3}.  Recently, this technique has been widely used in natural language processing. For example, BRAN~\cite{biaffine-1} and GCNN~\cite{biaffine-2} use a biaffine operator to score all mentions in parallel, while DGEDT~\cite{dgedt} introduces biaffine interactions between the representations of transformer and GCN for aspect level sentient classification task. However, how to incorporate biaffine in GCNs and how it performs remain to be explored.

\section{Biaffine Graph Convolutional Network}
Many existing GCNs adopting message passing mechanism have been proved to perform global smoothing across the whole graph~\cite{Ma2021CIKM}, however, the level of smoothness over different subgraphs can be different. More precisely, the nodes of the same class distribute across the non-contiguous subgraphs (as shown in Figure~1). In such cases, message passing may fail to propagate the class-relevant information from labeled set to unlabeled ones. Instead of capturing long-range relations between nodes with global diffusion, our method is aim to build direct connections between nodes and their long-range neighbors, which we refer to shortcuts. Towards this end, we employ biaffine attention mechanism to characterize the mutual correlations of a node and its high-order neighbors in terms of their local fields, on assumption that node features are locally smoothed over local fields (i.e., subgraphs). Our design is motivated by the conjecture that a node whose feature is similar to the local field of its distant neighbor may have strong correlation with this distant. neighbor, as the local field of a node is related the local structure of the node as well as its features.

\begin{figure*}[htpb]
	\centering
	\includegraphics[scale=0.6]{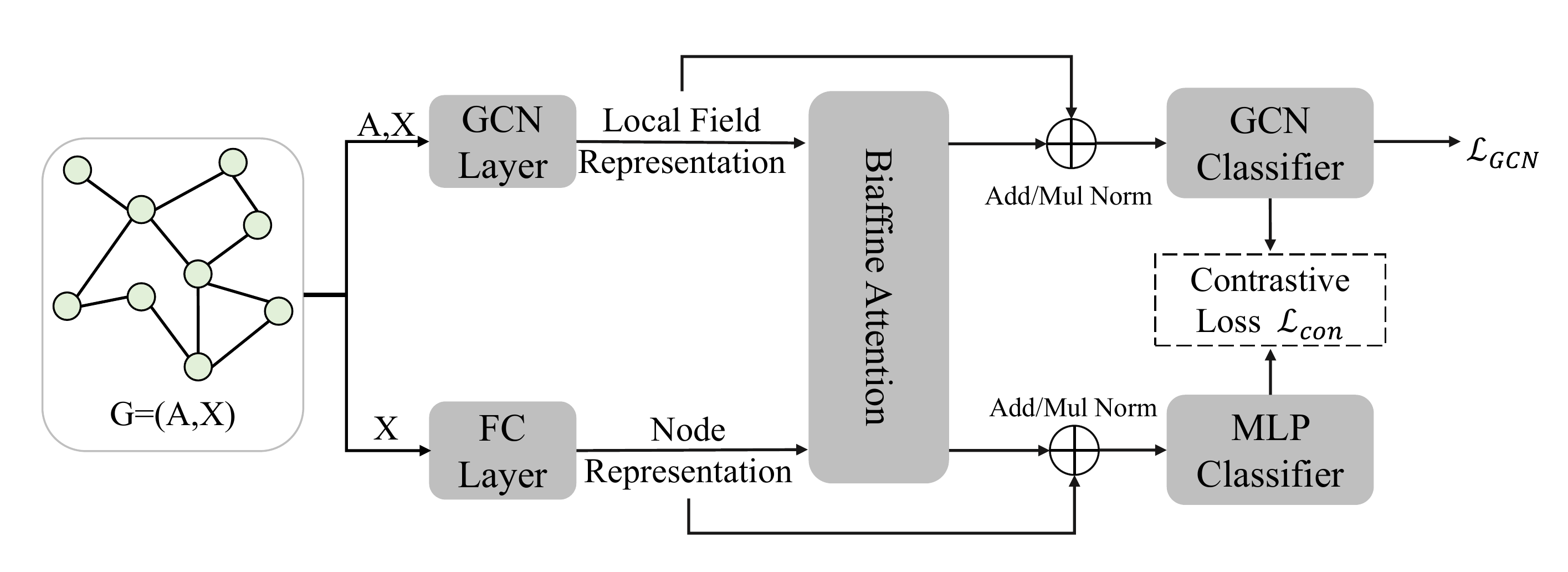}
	\caption{The architecture of BAGCN. Here the extremely simple configuration is used: each module (GCN and FC) only has one layer.}
	\label{fig:framework}
\end{figure*}

As shown in Figure~\ref{fig:framework}, our method consists of the following two components: (1) BAGCN layer that transforms the graph into two weighted dependency graphs with shortcuts; (2) multi-view contrastive learning for node classification by casting two dependency graphs as two views.

\subsection{BAGCN layer}
\label{sec:layer}
As aforementioned, the local field (e.g.subgraph) of a node can offer more fruitful information than a single node feature, so we opt for the first-order neighborhood subgraph as the local field of a node, whose representation is achieved using one-layer standard graph convolution:
\begin{equation}
	\boldsymbol{H_{c}} = ReLU(\boldsymbol{\widehat{A}H_c^{(0)}W_{1}} + \boldsymbol{\vec{b}_c})
\end{equation}
where $\boldsymbol{W_{1}} \in \mathbb{R}^{f \times d}$ and $\boldsymbol{\vec{b}_c} \in \mathbb{R}^{d}$ are learnable parameters. Note that $\boldsymbol{\widehat{A}}$ is a self-loop contained adjacency matrix, $\boldsymbol{H_{c}}\in \mathbb{R}^{n\times h}$ are thus the representations of the 1-hop subgraphs centered at each node. Without loss of generality, more higher order local field can be obtained in a similar way. Before scoring the dependency of a node on others using biaffine, we offer a feature transformation for node feature (i.e., attributes) with a fully-connected layer:
\begin{equation}
	\boldsymbol{H_{ego}} = ReLU(\boldsymbol{H_{ego}^{(0)}\theta} + \boldsymbol{\vec{b}_{\theta}})
\end{equation}
where $\boldsymbol{\theta} \in \mathbb{R}^{f \times d}$ and $\boldsymbol{\vec{b}}_{\theta} \in \mathbb{R}^{d}$ is the model parameter to be learned and $\boldsymbol{H_{ego}} \in \mathbb{R}^{n\times f}$. In particular, $\boldsymbol{H_{c}^{(0)}}=\boldsymbol{H_{ego}^{(0)}}=\boldsymbol{X}$.

Next, we use biaffine transformation to capture the mutual dependencies between $\boldsymbol{H_c}$ and $\boldsymbol{H_{ego}}$,
\label{sec:biaffine}

\begin{equation}\label{eq:affine_1}
	\boldsymbol{S_{1}} = softmax(\boldsymbol{H_{c}M_{1}H_{ego}^{T}}),
\end{equation}
\begin{equation}\label{eq:affine_2}
	\boldsymbol{S_{2}} = softmax(\boldsymbol{H_{ego}M_{2}H_{c}^{T}}),
\end{equation}
where $\boldsymbol{M_{1}}, \boldsymbol{M_{2}} \in \mathbb{R}^{d \times d}$ are the metric spaces to learn. $\boldsymbol{S_1}$ and $\boldsymbol{S_2}$ can be regarded as the constructed directed connections (including long-range dependencies) between nodes, whose weights reflect the degree of alignment of $\boldsymbol{H_c}$ and $\boldsymbol{H_{ego}}$ in metric spaces. In particular, $\boldsymbol{S_1}$ corresponds to the dependencies of ego nodes on the local fields of other nodes, whose values are the attention scores when taking ego-node features as queries and local field representations $\boldsymbol{H_c}$ as keys. Similarly, $\boldsymbol{S_2}$ depicts the dependencies of the local fields of the nodes on ego nodes. Those learned dependency relations allow the message passing directly to long-range or high order neighbors, which we simply implement it as follows:
\begin{equation}\label{eq:affine_3}
	\boldsymbol{H_{c}^{a}} = \boldsymbol{S_{1}H_{ego}}
\end{equation}
\begin{equation}\label{eq:affine_4}
	\boldsymbol{H_{ego}^{a}} = \boldsymbol{S_{2}H_{c}}
\end{equation}
where $\boldsymbol{H_{c}^{a}}$ is the result of affine projection from $\boldsymbol{H_{ego}}$ to $\boldsymbol{H_c}$, while $\boldsymbol{H_{ego}^{a}}$ reflects the affine from $\boldsymbol{H_{c}}$ to $\boldsymbol{H_{ego}}$.   


The above representations on top of biaffine-built direct interactions (i.e., $\boldsymbol{S_1}$ and $\boldsymbol{S_2}$) are further fused with the original features respectively as the final outputs of the BAGCN layer. Here we consider two fusion methods, namely, sum pooling and element-wise product,
\begin{equation}
\begin{split}
\label{eq:affine_gcnmix}
	\boldsymbol{{H_c}'} = Norm(\boldsymbol{H_c} + \boldsymbol{H_{c}^{a}}) \\ OR \quad \boldsymbol{{H_c}'} = Norm(\boldsymbol{H_c} \odot \boldsymbol{H_{c}^{a}})
\end{split}
\end{equation}
\begin{equation}
\begin{split}
\label{eq:affine_mlpmix}
	\boldsymbol{{H_{ego}}'} = Norm(\boldsymbol{H_{ego}} + \boldsymbol{H_{ego}^{a}}) \\ OR \quad \boldsymbol{{H_{ego}}'} = Norm(\boldsymbol{H_{ego}} \odot \boldsymbol{H_{ego}^{a}})
\end{split}
\end{equation}
where $\odot$ denotes the Hadamard product operation. We regard $\boldsymbol{{H_c}'}$ and $\boldsymbol{{H_{ego}}'}$ as two views of the node representations of the graph, with which we will use multi-view contrastive learning for node classification.

\noindent\textbf{Biaffine Mapping vs. GAT attention.}
Note that biaffine graph convolutional layer can be viewed as a type of attention mechanism, it is interesting to compare it with the classical graph attention model GAT. The first difference is that GAT leads to the ego to ego attention at the first convolution layer and the local field  to local field attention at higher layers, while our biaffine graph convolution allows the mutual attention between ego and local field (i.e.,subgraph). Moreover, GAT is a local self-attention, whose goal is to assign weights to the first-order neighbors. In contrast, biaffine attention globally learns correlations of the nodes, thus establishing shortcuts between nodes that are not connected directly. Therefore, BAGCN needs to learn $2\times n^{2}$ attention scores, a bit more than that of GAT (i.e., $2\|E\|$).


Next, we show that BAGCN layer is able to capture the features of higher-order neighbors that GCN cannot. Given a graph \bm{$G$} whose symmetric normalized Laplacian matrix is \bm{$\widehat{A}$}. In particular, there is no connection between nodes $i$ and $j$, if $\boldsymbol{\widehat{A}}_{i,j}=0$.  For illustrative purpose, we consider $\boldsymbol{W_{1}}=\boldsymbol{W_{2}}=\boldsymbol{I}$ in the equations~(\ref{eq:affine_1}) and~(\ref{eq:affine_2}), then we have:
\begin{equation}\label{eq:peoposition1}
	\boldsymbol{S_1} = softmax(\boldsymbol{H_{c}H_{ego}^{T}}), \boldsymbol{H_{c}^{a}} = \boldsymbol{S_1 H_{ego}}
\end{equation}
In general, $\boldsymbol{S_1}$ is a dense matrix, and if the vectors in $\boldsymbol{H_{c}}$ and $\boldsymbol{H_{ego}}$ are not orthogonal, $\boldsymbol{S_{1}}(i,j)$ is a non-zero real value. Let $m_{i}  = \sum _{j \in \mathcal{N}_{i}}\mathbb{I}(\boldsymbol{S_{1}}(i,j))$ be the number of nodes that can pass information to node $i$ on top of $\boldsymbol{S_{1}}$, where $\mathbb{I}(\cdot )$ is the indicator function that takes $1$ when $\boldsymbol{S_{1}}(i,j) \neq 0$, otherwise 0.


Considering that two or three convolution layers for GCN are generally optimal in practical settings, we compare BAGCN convolution (i.e., Eq.~\ref{eq:peoposition1}) with the simplified two-layer GCN without consideration of parameters described as follows:
\begin{equation}\label{eq:peoposition2}
	\boldsymbol{H^{GCN}} = \boldsymbol{\widehat{A}}(\boldsymbol{\widehat{A}H_{ego}}) = \boldsymbol{\widehat{A}^2H_{ego}}
\end{equation}
where $\boldsymbol{\widehat{A}^2}$ depicts the 2-hop neighbors of the nodes. In this setting, the number of nodes from which node $i$ can aggregate information is ${m_{i}}'= \sum _{j \in \mathcal{N}_{i}^{2-hop}} \mathbb{I}(\boldsymbol{\widehat{A}}^2_{i,j})$ nodes. Recall that $\boldsymbol{S_1}$ is generally dense , i.e., $m_i \rightarrow n$, while the expected $m'$ for all nodes in the vanilla GCN can be found $\overline{{m}'} < \frac{|E|^{2}}{n^{2}} << n \approx m_{i}$. 
Therefore, one-layer BAGCN can capture more information from higher-order nodes without deep architectures.

\subsection{Multi-view contrastive learning}
\label{sec:loss}
To exploit the feature representations from two types of dependency relations (i.e., $\boldsymbol{S_1}$ and $\boldsymbol{S_2}$), a common practice is to combine them with concatenation or other pooling techniques. Instead, here we adopt multi-view contrastive learning scheme to enhance the prediction performance. Specifically, the outputs of BAGCN module $(\boldsymbol{H_{c}'}, \boldsymbol{H_{ego}'})$ are fed to two different predictors respectively. For $\boldsymbol{H_{c}'}$ of one view, we first use one-layer graph convolution to update context information on top of learned dependencies, and then map the resultant features to the classification dimension, which is formulated as:
\begin{equation}
	\boldsymbol{\widehat{Y}_{gcn}} = \boldsymbol{(\widehat{A}{H_{c}}')W_{c}}
\end{equation}
where $\boldsymbol{W_{c}} \in \mathbb{R}^{d \times C}$ and $\boldsymbol{\widehat{Y}_{gcn}}$ is the labeling matrix whose cell $\widehat{Y}_{gcn}(i,j)$ corresponds to the probability of node $i$ belonging to label $j$. For $\boldsymbol{H_{ego}'}$ of another view, as it is the projection of node features on the contextual representation space, we directly feed it to a multi-layer perceptron (MLP) classifier:
\begin{equation}
	\boldsymbol{\widehat{Y}_{fc}} = f_{MLP}(\boldsymbol{{H_{ego}}'}, \boldsymbol{\theta_{c}})
\end{equation}
\par
\noindent\textbf{Contrastive learning.}
In order to learn rich information about nodes, we use $L_2$ regularization to enforce the prediction results between two views to be similar to each other. We define the objective as follows:
\begin{equation}\label{eq:consis_loss}
	\mathcal{L}_{con} = \frac{1}{2}(\left \| \boldsymbol{\overline{Y}} - \boldsymbol{\widehat{Y}_{gcn}}\right \|_{2}^{2} + \left \|\boldsymbol{\overline{Y}} - \boldsymbol{\widehat{Y}_{fc}} \right \|_{2}^{2})
\end{equation}
where $\boldsymbol{\overline{Y}}$ is the average over two views' predicted logits, i.e., $\boldsymbol{\overline{Y}} = \frac{1}{2}(\boldsymbol{\hat{Y}_{gcn}} + \boldsymbol{\hat{Y}_{fc}})$.
Considering the distribution of predicted labels may be very flat, we utilize the \emph{sharpening}~\cite{sharp} approach to reduce the uncertainty of the averaged label prediction:
\begin{equation}\label{eq:sharp}
	\boldsymbol{\overline{Y}}_{i,q} =  \frac{\boldsymbol{\overline{Y}}_{i,q}^{\frac{1}{\mathcal{T}}}}{\sum_{c=0}^{C-1}\boldsymbol{\overline{Y}}_{i,c}^{\frac{1}{\mathcal{T}}}},(0\leq q \leq C-1)
\end{equation}
where the hyperparameter $\mathcal{T}$ serves as the "temperature" adjuster for the categorical distribution~\cite{deeplearning}. In particular, as $\mathcal{T}\rightarrow 0$, the logits will be sharpened into a one-hot vector. On the other hand, $\mathcal{T}\rightarrow \infty $ implies that the logits of all classes will become equal, which is not we expect. In general, for $\mathcal{T}>1$, Eq.~\ref{eq:sharp} acts as passivation. 
By constrastively learning node labels, feature representations of the nodes can be mediated between two views. It should be noted that, differing from directly minimizing the prediction errors between two views, i.e., $\mathcal{L}_{con} = \left \| \boldsymbol{\widehat{Y}_{gcn}} - \boldsymbol{\widehat{Y}_{fc}}\right \|_{2}^{2}$ , Eq.~\ref{eq:consis_loss} allows the two views to align with their average prediction, which empirically show consistent outperformance on node classification benchmarks (see section 4.5).
\par
\noindent\textbf{Classification.}
Although there are two classifiers (i.e., one-layer GCN and one-layer MLP) for the same task, MLP may be inferior to one-layer GCN predictor. This is attributed to the input of MLP, which is the combination of the original node features and affine representations, less informative than local field. So we choose to use the one-layer GCN classifier for inference, whose classification training loss reads as
\begin{equation}\label{eq:loss_1}
	\mathcal{L}_{GCN} = -\frac{1}{C}\sum_{i=1}^{C} \boldsymbol{Y^{T}_{i}}\log\boldsymbol{\widehat{Y}_{gcn_{i}}}
\end{equation}
where $Y$ is the ground-truth.
Combining $\mathcal{L}_{GCN}$ with contrastive loss, the total loss of BAGCN is then as follows:
\begin{equation}\label{eq:final_loss}
	\mathcal{L} = \mathcal{L}_{GCN} + \lambda \mathcal{L}_{con}
\end{equation}
where $\lambda$ is a hyper-parameter that trade off between the two losses.

\noindent\textbf{Complexity and Limitations.}
Computing one-layer GCN requires $\mathcal{O}(|E|fd)$ time complexity, where $d$ denotes the number of hidden units, while the complexity of FC layer is $\mathcal{O}(nfd)$, and the implementation  of biaffine mapping takes $\mathcal{O}(nd(n+d))$ time complexity, so running the BAGCN layer takes $\mathcal{O}(|E|fd+ nd(f+n+d))$ computational time, which is less efficient than running GCN. In particular, a big challenge that biaffine's global attention brings is the scalability when faced with very large graphs, which we include in our future work.

\section{Experimental Results}

To demonstrate the effectiveness of the designed architecture, we ran a set of semi-supervised node classification tasks to compare the proposed BAGCN method with multiple competitive GCN models. In addition, we test the robustness of BAGCN against adversarial attacks and answer whether it is able to capture higher-order information when quite a few training labels are available.
\subsection{Experiment Setup}
\textbf{Datasets.} We use nine benchmark datasets widely used in node classification tasks in the literature~\cite{2016citedatasets,kipf2017gcn, graphSAGE2017,gcnII,grand}, including three standard citation networks~\cite{2016citedatasets,kipf2017gcn}, namely, Cora, Citeseer and Pubmed, and other larger graph datasets~\cite{datasetscitefull,datasetcoauther}. Table~\ref{tb:dataintroduce} summarizes the statistics of the those benchmark datasets.
\begin{itemize}
	\item \emph{Citation networks} include Cora, Citeseer and Pubmed. They are composed of papers as nodes and their relationships such as citation relationships, common authoring. Node feature is a one-hot vector that indicates whether a word is present in that paper. Words with frequency less than 10 are removed.
	
	\item \emph{cora-ML}  Like cora, this graph is also extracted from the original data of the entire network.
	
	\item \emph{Photo} is segment of the Amazon co-purchase graph, where nodes represent goods, edges indicate that two goods are frequently bought together, node features are bag-of-words encoded product reviews, and class labels are given by the product category.
	
	\item \emph{Coauthor CS and Coauthor Physics} are co-authorship graphs based on the Microsoft Academic Graph from the KDD Cup 2016 challenge 3. Here, nodes are authors, that are connected by an edge if they co-authored a paper; node features represent paper keywords for each author’s papers, and class labels indicate most active fields of study for each author.
	
	\item \emph{DBLP} is a citation network dataset. The citation data is extracted from DBLP, ACM, MAG (Microsoft Academic Graph), and other sources. Each paper is associated with abstract, authors, year, venue, and title.
\end{itemize}
Following the setting of prior work~\cite{2016citedatasets}, we apply the standard fixed training/validation/testing split on Cora, Citeseer and Pubmed. For the rest of the datasets, we utilize 20 labeled nodes per class for training, 30 labeled nodes for validation, and the rest for testing set.

 \begin{table*}[htbp]
 \small
 	\centering
 	\caption{The statistics of the datasets}
 	\label{tb:dataintroduce}
 	\begin{tabular}{cccccc}
 		\toprule
 		\textbf{Datasets} & \textbf{Nodes} & \textbf{Edges} & \textbf{Train/Valid/Test Nodes} & \textbf{Features} & \textbf{Classes} \\
 		\midrule
 		Cora & 2,708 & 5,429 & 140/500/1000 & 1,433 & 7 \\
 		Citeseer & 3,327 & 4,732 & 120/500/1,000 & 3,703 & 6 \\
 		Pubmed & 19,717 & 44,338 & 60/500/1,000 & 500 & 3 \\
 		Cora-ML & 2,995 & 16,316 & 140/210/2,545 & 2,879 & 7 \\
 		Photo & 7,487 & 119,043 & 160/240/7,084 & 745 & 8 \\
 		Coauthor CS & 18,333 & 81,894 & 300/450/17,583& 6,805 & 15 \\
 		Coauthor Physics & 34,493 & 247,962 & 360/540/33,593 & 100 & 18 \\
 		DBLP & 17,716 & 105,734 & 80/120/17,516 & 1,639 & 4 \\
 		Cora-Full & 19,793 & 126,842 & 1,400/2,100/15,293 & 8,710 & 70 \\
 		\bottomrule
 	\end{tabular}
 \end{table*}

\textbf{Baseline.} We include several shallow models: APPNP~\cite{appnp}, graphSAGE~\cite{graphSAGE2017}, BGCN~\cite{bgcn}~\footnote{We report the best results of the two variants of BGCN.}, MixHop~\cite{mixhop2019},DAGCN~\cite{dagnn}, and deep network models: JKNet~\cite{jknet},  DGI~\cite{dgi}, GCNII~\cite{gcnII}, many of which are are proposed quite recently with promising performance. We also compare with classical GCN~\cite{kipf2017gcn} and GAT~\cite{2018gat}. Besides, we specially include Mixture model network (MoNet)~\cite{monet} and all three variants of GraphSAGE for large graph datasets. Among these baselines, DAGNN, JK-net, APPNP and GCNII are designed with the ability of capturing long-distance nodes' information. Further more, we compare our method with recent over-squashing oriented approaches, namely, +FA method~\cite{oversquashing} and Stochastic discrete ricci flow (SDRF)~\cite{oversquashing22}.
\begin{itemize}
	\item \textbf{GCN} uses the normalized adjacency matrix as the weight of message passing and achieves the good performance in practice when two layers are adopted. So we categorize it into shallow model.
	
	\item \textbf{GAT} calculates the attention of the connected nodes, and propagates the attention value as the weight of the message.
	
	\item \textbf{GraphSAGE}  is an inductive learning framework that uses the attribute information of nodes to efficiently generate the feature representation of unknown nodes.
	
	\item \textbf{Mo-net}  adopts a different approach to assign different weights to a node’s neighbors.
	
	\item \textbf{DGI} is an unsupervised learning method that obtains node embeddings by maximizing local and global mutual information.
	
	\item \textbf{JK-Net} aggregates different domains by adaptively learning nodes at different locations, thereby improving the representation of nodes.
	
	\item \textbf{APPNP}  uses the relationship between graph convolution network (GCN) and PageRank to derive an improved propagation scheme based on personalized PageRank.
	
	\item \textbf{GRACE} generates two graph views by removing edges (RE) and masking features (MF), and learns the representation by maximizing the node representation of the two views.
	
	\item \textbf{DAGNN} use an adaptive adjustment mechanism, so that the information of each node’s local and global neighbors can be adaptively balanced, leading to a more discriminating node representation.
	
	\item \textbf{GCNII}  is a powerful deep GCN, which is a GCN with initial residual connection and identity mapping.
	
	\item \textbf{+FA}  is to alleviate the over-squashing problem by changing the last layer of GCN into a fully connected layer.
	
	\item \textbf{SDRF} The main idea of SDRF(Stochastic Discrete Ricci Flow) is to remove the edges of negative curvature and increase the links between nodes.
\end{itemize}


\textbf{Experimental details.} For large-scale datasets, most of the baseline models are performed under the same data splitting as the original papers, so the parameters are set following the original ones. For the classic graph benchmarks Cora, Citeseer, and Pubmed, we search the parameters of the models, and use the settings corresponding to the best performance of the models. The results of some models shown here exceed the reported in the original papers.

For our model BAGCN, besides biaffine mapping, there are only two graph convolution layers involved: one layer for local field representation, and one layer for classification. We use Batch normalization~\cite{batchnorm} for every hidden layer and set the initial learning rate of the Adam optimizer~\cite{adam} to 0.01. The number of hidden units are searched in $\{32, 64, 128\}$ and the dropout ratio is searched in $\{0.1, 0.2, 0.3, 0.4, 0.5, 0.6\}$ for each dense connected layer. We also apply $L_2$ norm regularization and set weight decay to 5e-4 for each layers; Other hyperparameter settings are: 200 epochs for running models; 0.7 for temperature $\mathcal{T}$ and $\lambda$ = 1 in penalizing multi-view contrastive learning loss. All experiments are implemented in PyTorch on 2 NVIDIA RTX3090 24G GPUs with CUDA 11.0.

\begin{table*}[htbp]
\small
	\centering
	\caption{Classification accuracy (\%) on 5 large datasets. Results are averaged over 20 runs. The best performance for each dataset is highlighted in bold
    and the second best performance is underlined for comparison.}
	\label{tb:results_large}
	\begin{tabular}{cccccc}
		\toprule
		\textbf{Datasets} & \textbf{Photo} & \textbf{Coauthor CS} & \textbf{Coauthor Physics} & \textbf{DBLP} & \textbf{Cora-Full} \\
		\midrule
		MLP & 69.64\scriptsize $\pm$3.8 & 88.30\scriptsize $\pm$0.7 & 88.91\scriptsize $\pm$1.1 & 48.33\scriptsize $\pm$2.8 & 18.66\scriptsize $\pm$1.0 \\
		GCN & 91.21\scriptsize $\pm$1.2 & 91.13\scriptsize $\pm$1.5 & 92.78\scriptsize $\pm$1.0 & 83.86\scriptsize $\pm$1.9 & 11.63\scriptsize $\pm$1.4  \\
		GAT & 85.72\scriptsize $\pm$20.3 & 90.49\scriptsize $\pm$0.6 & 92.53\scriptsize $\pm$0.9 & 82.99\scriptsize $\pm$2.2 & 31.40\scriptsize $\pm$3.5  \\
		MoNet & 91.24\scriptsize $\pm$1.3 & 90.82\scriptsize $\pm$1.6 & 92.46\scriptsize $\pm$0.9 & 83.28\scriptsize $\pm$1.8 & 28.78\scriptsize $\pm$2.2  \\
		GraphSAGE-mean & 91.43\scriptsize $\pm$1.3 & 91.32\scriptsize $\pm$2.9 & 93.06\scriptsize $\pm$0.8 & 83.98\scriptsize $\pm$1.4 & 35.48\scriptsize $\pm$1.7  \\
		GraphSAGE-maxpool & 90.42\scriptsize $\pm$1.3 & 85.04\scriptsize $\pm$1.1 & 90.26\scriptsize $\pm$1.2 & 82.66\scriptsize $\pm$2.4 & 34.15\scriptsize $\pm$2.0 \\
		GraphSAGE-meanpool & 90.72\scriptsize $\pm$1.6 & 89.66\scriptsize $\pm$0.9 & 92.65\scriptsize $\pm$1.0 & 83.55\scriptsize $\pm$1.8 & 35.12\scriptsize1.7 $\pm$\\
		GRACE & 91.46\scriptsize $\pm$0.28 & \underline{92.53\scriptsize $\pm$0.1} & 95.63\scriptsize $\pm$0.1 & \underline{84.16\scriptsize $\pm$0.1} & 46.37\scriptsize $\pm$0.5 \\
		DAGNN & \underline{92.02\scriptsize $\pm$0.8} & \textbf{92.78\scriptsize $\pm$0.9} & 94.01\scriptsize $\pm$0.6 & 79.49\scriptsize $\pm$1.1 & 30.01\scriptsize $\pm$1.5 \\
		GCNII & 88.48\scriptsize $\pm$2.1 & 92.41\scriptsize $\pm$0.5 & 93.94\scriptsize $\pm$0.4 & 76.52\scriptsize $\pm$1.9 & \underline{50.16\scriptsize $\pm$1.4} \\
		\midrule
		BAGCN-mul & 90.71\scriptsize $\pm$1.1 & 91.48\scriptsize $\pm$0.9 & \textbf{96.08\scriptsize $\pm$0.5} & 83.10\scriptsize $\pm$3.5 & 48.56\scriptsize $\pm$3.5 \\
		BAGCN-add & \textbf{92.35\scriptsize $\pm$0.9} & 92.27\scriptsize $\pm$0.5 & \underline{95.88\scriptsize $\pm$0.8} & \textbf{85.47\scriptsize $\pm$3.4} & \textbf{52.67\scriptsize $\pm$2.1} \\
		\bottomrule
	\end{tabular}
\end{table*}

\subsection{Comparison with State-of-the-Art}

\begin{table}[htbp]
\small
	\centering
	\caption{Classification accuracy (\%) on 4 widely-used benchmarks. Results are averaged over 20 runs.}
	\label{tb:results_small}
	\begin{tabular}{cccccc}
		\toprule
		\textbf{Datasets} & \textbf{Cora} & \textbf{Citeseer} & \textbf{Pubmed} & \textbf{Cora-ML} \\
		\midrule
		MLP & 61.6\scriptsize $\pm$0.6 & 61.0\scriptsize $\pm$1.0 & 74.2 \scriptsize $\pm$ 0.7 & 60.9\scriptsize $\pm$1.1  \\
		GCN & 81.7\scriptsize $\pm$0.4 & 70.9\scriptsize $\pm$0.5 & 78.8\scriptsize $\pm$ 0.6 & 82.2\scriptsize $\pm$1.5  \\
		GAT & 83.0\scriptsize $\pm$0.7 & 72.5\scriptsize $\pm$0.6 & 79.1\scriptsize $\pm$0.4 & 82.2\scriptsize $\pm$1.9  \\
		DGI & 82.5\scriptsize $\pm$0.7 & 71.6\scriptsize $\pm$0.7 & 78.4\scriptsize $\pm$0.7 & 82.4\scriptsize $\pm$1.2 \\
		JKNet & 81.3\scriptsize $\pm$0.6 & 70.4\scriptsize $\pm$1.1 & 78.1\scriptsize $\pm$0.4 & 81.1\scriptsize $\pm$1.2 \\
		BGCN & 82.0\scriptsize $\pm$0.1 & 71.9\scriptsize $\pm$0.0 & 79.4\scriptsize $\pm$0.1 & 82.1\scriptsize $\pm$0.4 \\
		GRACE & 83.3\scriptsize $\pm$0.5 & 72.1\scriptsize $\pm$0.5 & 79.0\scriptsize $\pm$0.2 & \underline{83.3\scriptsize $\pm$0.3} \\
		MixHop & 81.9\scriptsize $\pm$0.4 & 71.4\scriptsize $\pm$0.8 & \textbf{80.8\scriptsize $\pm$0.6} & 81.7\scriptsize $\pm$1.5 \\
		APPNP & 83.3\scriptsize $\pm$0.5 & 71.7\scriptsize $\pm$0.6 & \underline{80.1\scriptsize} $\pm$0.2 &  82.1\scriptsize $\pm$1.1 \\
		\midrule
		GCN+FA & 81.6 \scriptsize $\pm$ 0.2 & 70.5\scriptsize $\pm$0.2 & 79.5 \scriptsize $\pm$ 0.1 & 82.3 \scriptsize $\pm$ 1.1 \\
		SDRF & 82.8\scriptsize $\pm$0.2 & 72.6\scriptsize $\pm$0.2 & 79.1\scriptsize $\pm$0.1 & 82.5 \scriptsize $\pm$ 1.4\\
		\midrule
		BAGCN-mul & \textbf{83.7\scriptsize $\pm$0.2} & \underline{72.6\scriptsize $\pm$0.1} & 78.5\scriptsize $\pm$0.1 & 82.7\scriptsize $\pm$1.7 \\
		BAGCN-add & \underline{83.3\scriptsize $\pm$0.2} & \textbf{73.0\scriptsize $\pm$0.7} & 78.5\scriptsize $\pm$0.1 & \textbf{83.9\scriptsize $\pm$1.3} \\
		\bottomrule
	\end{tabular}
\end{table}
Table~\ref{tb:results_large} reports the semi-supervised node classification performance on large datasets. We take the results of the baseline methods from original work if available (e.g., GRACE~\cite{grace} and DAGNN~\cite{dagnn}). All results of the proposed BAGCN are averaged over 20 runs with random weight initializations on random training/validation/test splits, where the class distribution is guaranteed to be uniform in the training set. The results reported in Table~\ref{tb:results_large} show that BAGCN achieves the state-of-the-art performance on large-scale benchmark datasets except Coauthor CS where our model is still competitive but slightly inferior to DAGCN and GRACE. Notably, GCN shows strong performance on large graphs, but our model improves the performance over GCN by a margin of 1.14\%, 1.14\%, 3.30\%, 1.61\%, and 41.04\% (absolute differences) on Photo, Coauthor CS, Coauthor Physics, DBLP and Cora-Full, respectively, which demonstrate the benefits of long-distance dependency learning. It is also worthwhile to note that two deep modesl (i.e., GCNII and  DAGCN) do not seem to offer any advantages over our shallow model.

We also compare BAGCN with deep models on citation datasets. The results are reported in Table~\ref{tb:results_small}. For fair comparisons, we use the fixed split, namely, 20 labeled nodes per class as the training set, 500 nodes as the validation set, and 1000 nodes as the test set for all models, following the previous work ~\cite{2016citedatasets}. For each model, we conduct 100 runs with 100 random seeds, which is commonly adopted to evaluate performance by the community. BAGCN model shows competitively strong performance compared to the baselines. On Cora, Citeseer, and Cora-ML, we observe 2.0\%, 2.1\%, and 1.5\%  absolute improvement over GCN. When compared to deep models (e.g., JKNet, DGI and APPNP), our model is distinctly superior to them. On the other hand, when compared with the shallow model MixHop that is also designed to learn higher-order neighborhood relations with one convolutional layer, BAGCN still outperforms it in 3 out of 4 benchmark datasets, e.g., at least 1.6\% absolute improvement over MixHop on Cora, Citeseer and Cora-ML. Finally, we compare the results with two over-squashing oriented approaches including GCN+FA and SDRF. The results shown in the last four lines of Table~\ref{tb:results_small} suggest the BAGCN also achieves leading performance in most datasets.

It is noteworthy that BAGCN achieves superior performance on the benchmark datasets with a wide range of size, compared to strong baselines and recent competing deep models, which verifies the effectiveness and expressiveness of our proposed model. For example, our shallow architecture can outperforms strong baseline APPNP with 10 layers. This suggests that the receptive fields of the nodes are successfully expanded by biaffine attention.
\begin{figure*}
	\centering
	{
	\subfigure[Original-Cora(id 1423)]{
	\hspace{-0.5cm}
	\includegraphics[scale=0.27]{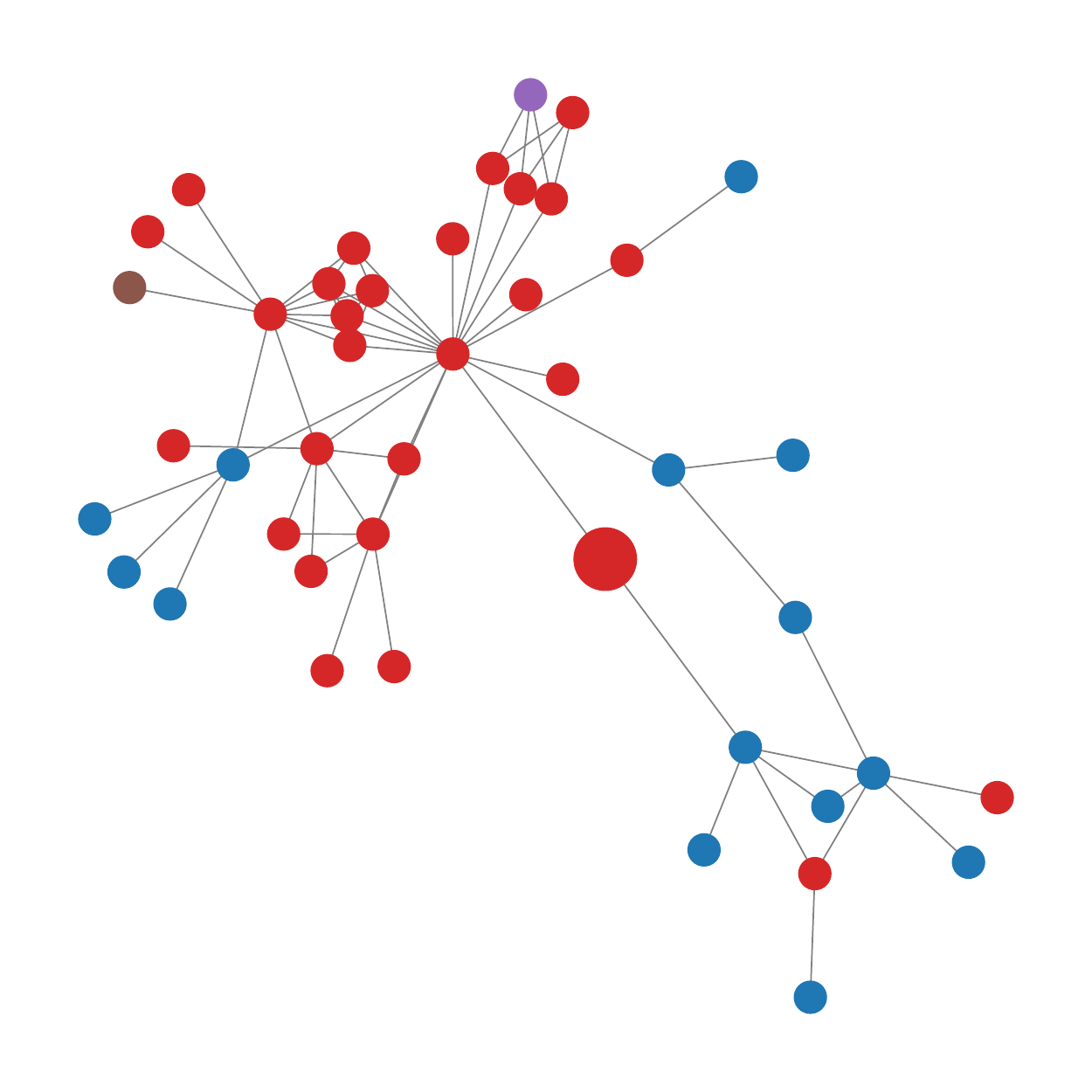} 
	}
	\subfigure[Original-Cora(id 2459)]{
	\includegraphics[scale=0.27]{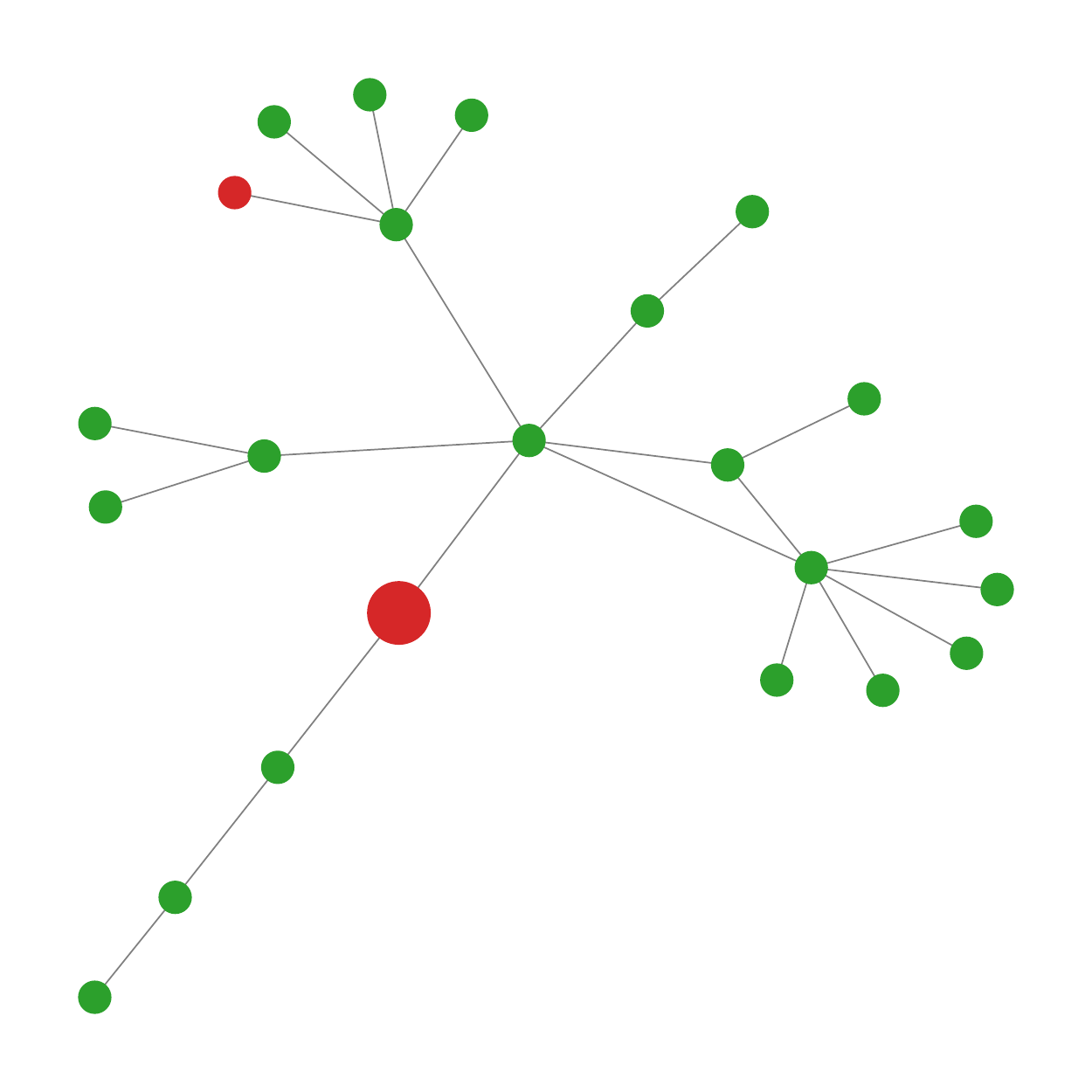}
	}
	\subfigure[Original-Citeseer(id 1022)]{
	\includegraphics[scale=0.27]{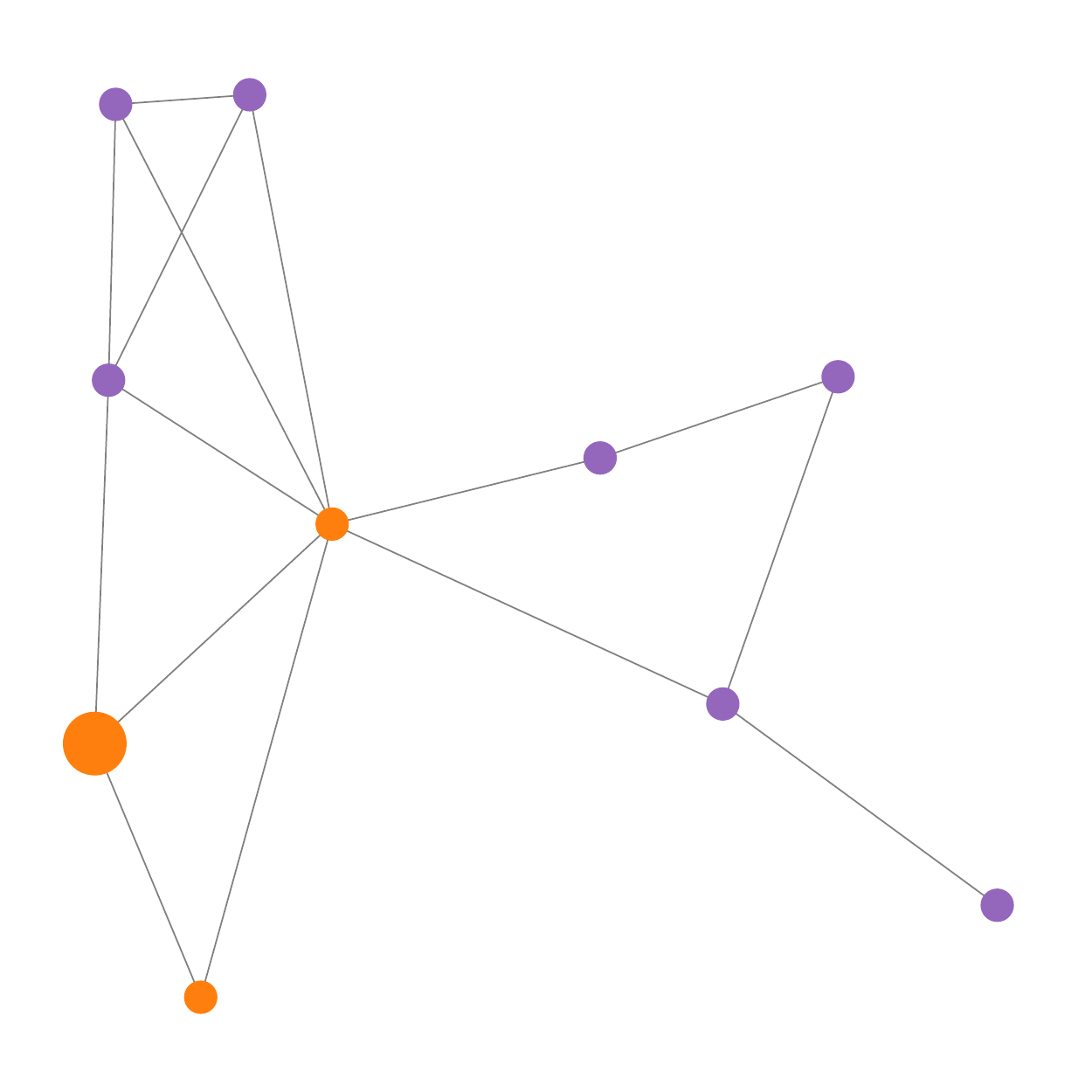}
	}
	\subfigure[Original-Citeseer(id 1423)]{
		\includegraphics[scale=0.27]{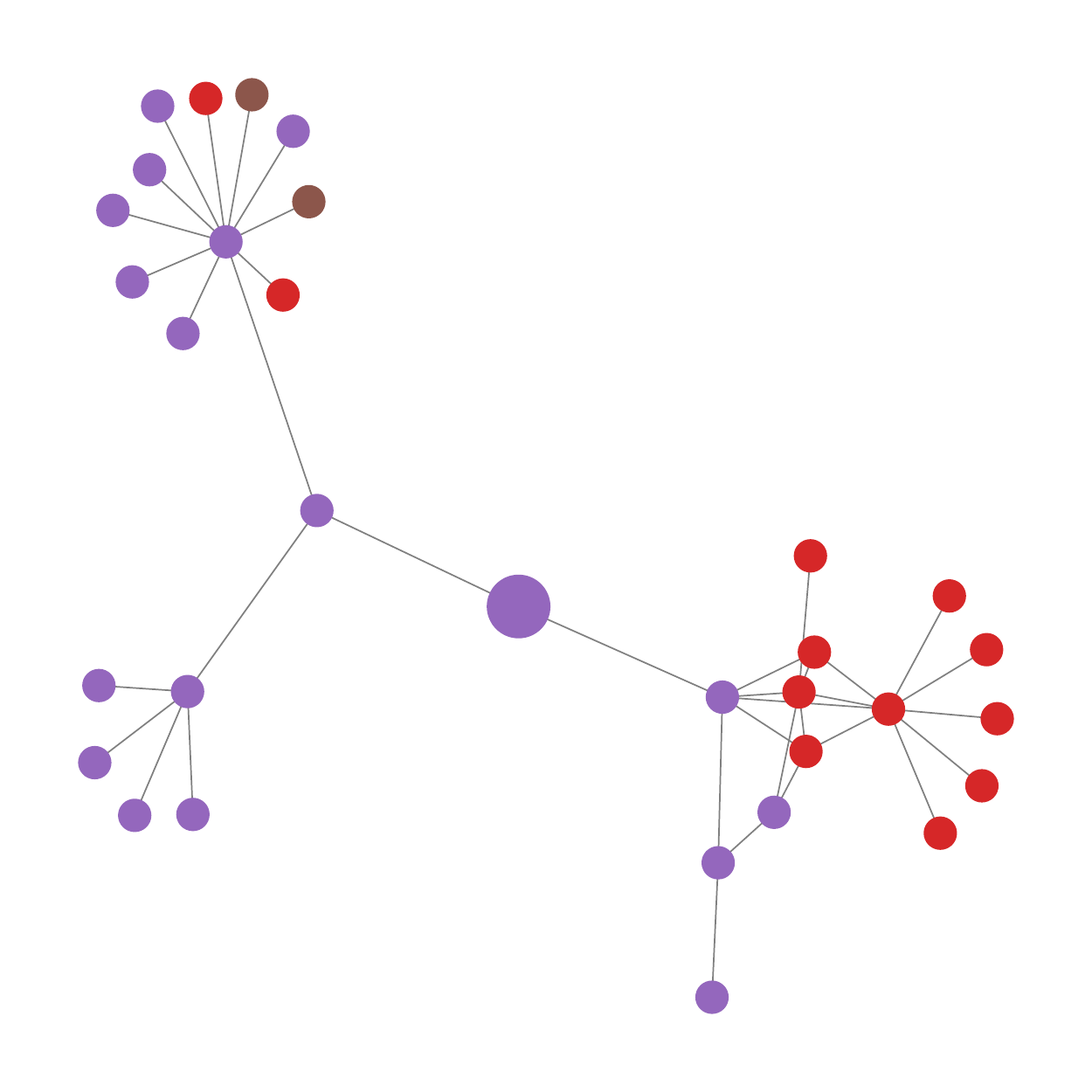}
		
	}
	\subfigure[BAGCN-Cora(id 1423)]{
	\hspace{-0.5cm}
		\includegraphics[scale=0.27]{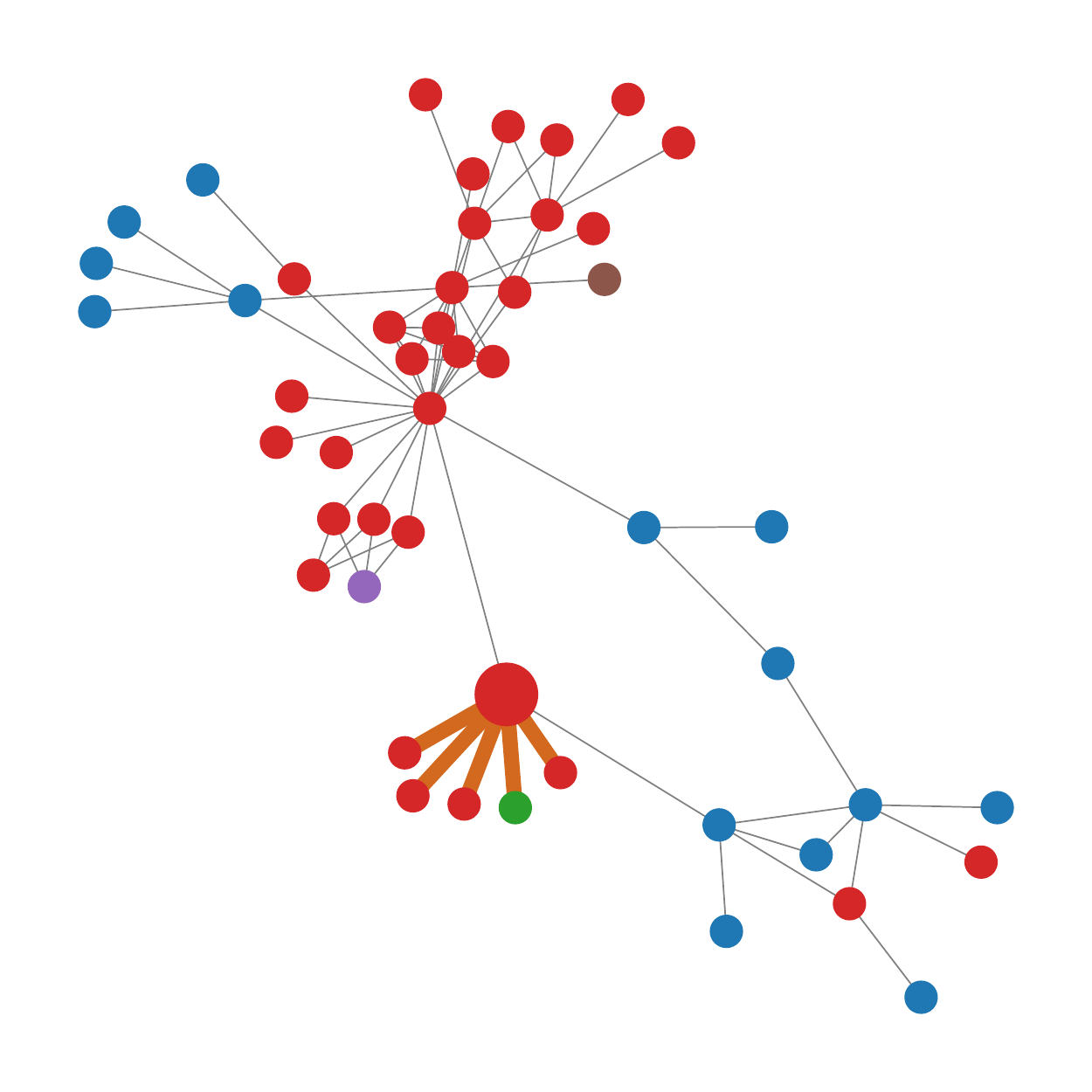}
		
	}
	\subfigure[BAGCN-Cora(id 2459)]{
		\includegraphics[scale=0.27, angle=90]{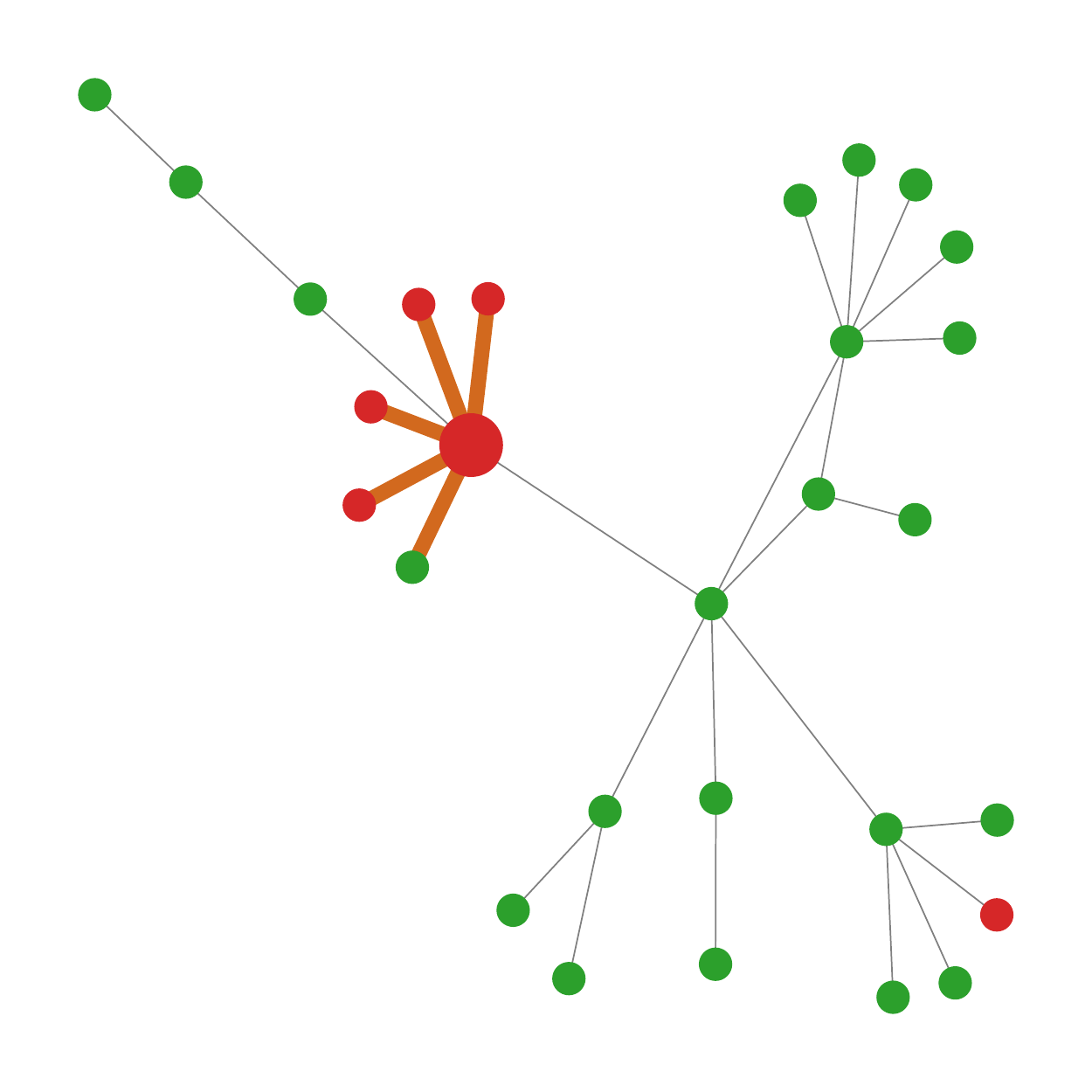}
	}
	\subfigure[BAGCN-Citeseer(id 1022)]{
		\includegraphics[scale=0.27]{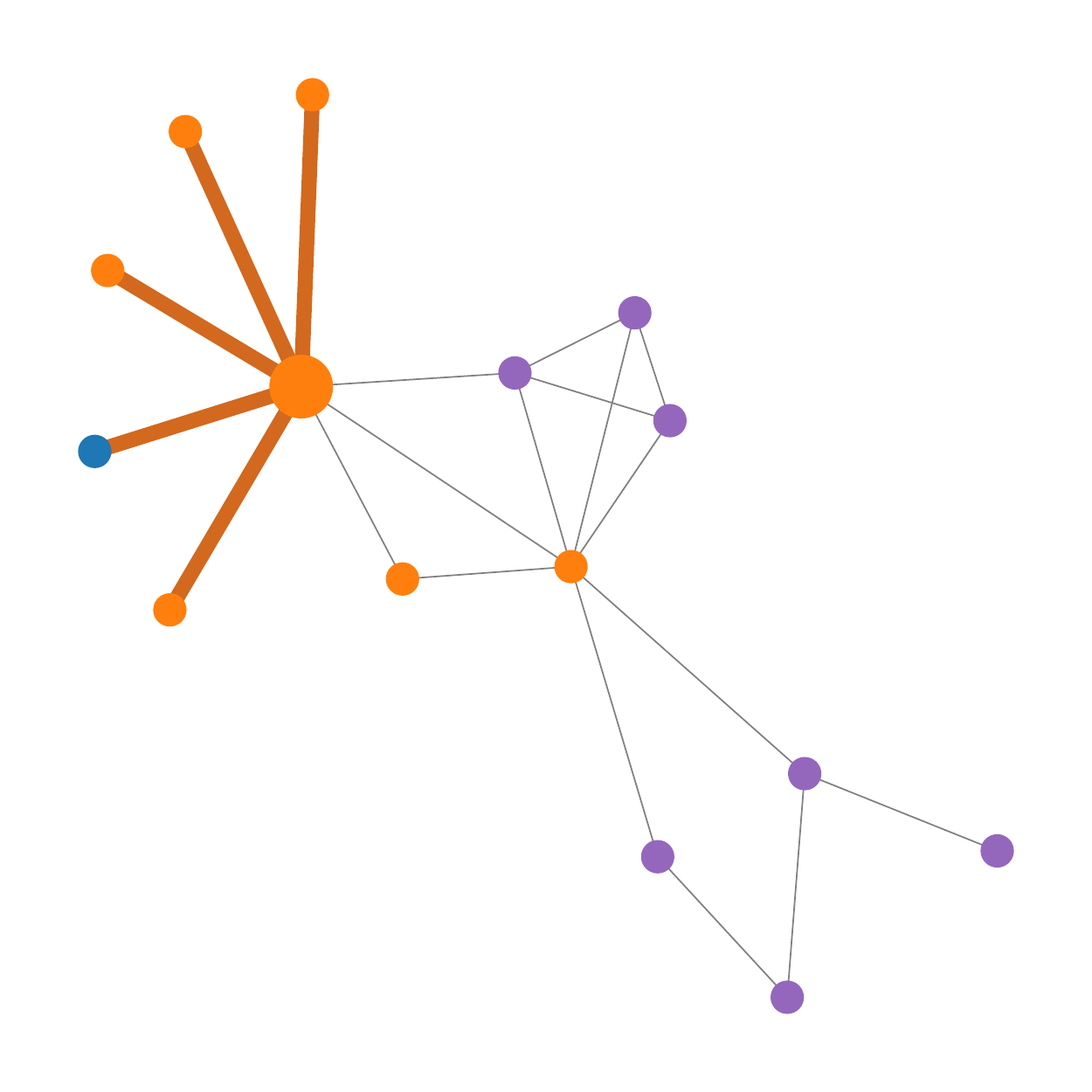}
	
	}
	\subfigure[BAGCN-Citeseer(id 1423)]{
		\includegraphics[scale=0.27]{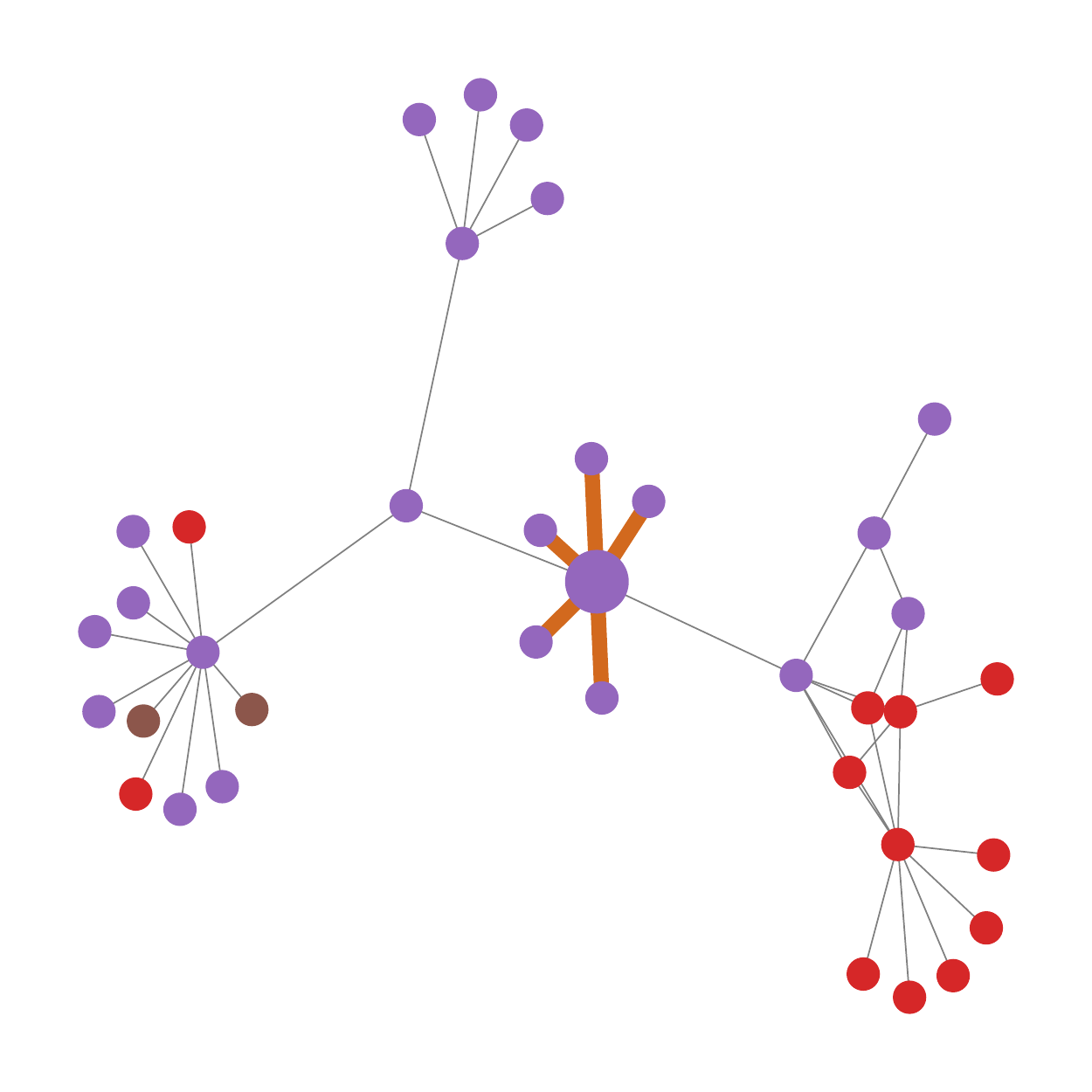}
		
	}
	\caption{Case studies of biaffine attention on two datasets (i.e., Cora and Citeseer) where different colors represents different class labels. The 3-hop ego-graphs of the central/ego nodes (bigger dots in the plots) are extracted from original graphs, corresponding to the first row in the figure. The biaffine-attentioned long-distance dependencies using BAGCN-add are highlighted in chocolate in each ego-graph.(bottom row)}
	\label{fig:networkx}
	}
\end{figure*}

\subsection{The effectiveness of biaffine on long-range dependency learning}
In this section, we evaluate the effectiveness of biaffine mapping on learning long-range dependency between nodes. We visualize the 3-hop ego-graph~\footnote{the $k$-hop ego-graph is the subgraph consisting of ego and its neighbors within $k$ hops.} structure of the nodes of interest in Figure~\ref{fig:networkx}, where different colors on nodes indicate different labels. Specifically, we select the top-k nodes for target nodes according to the values in node dependency matrix $S_1$ learned by BAGCN-add and add connections between those nodes and the targets in orange color. One observation is that the top-k nodes have no connections to the 3-hop subgraph (except for ego node), implying that the long-distance dependency outside of 3-hop neighborhood is captured. Moreover, almost all long-distance dependent nodes have the same class label as the egos, which increases the local homophily~\footnote{The local homophily of a node is defined as its tendency to connect to the nodes of the same type.} of the target nodes and thus facilitates the node classification. The above observations suggest that biaffine attention is able to build direct dependency between nodes and their long-range neighbors.

It is also noteworthy that not all nodes obtain long-distance relationships via biaffine attention. In effect, we find some nodes that have no long-range connections exhibit the following properties: their degrees are generally larger than the average degree of the graph, and they tend to have greater relative degree compared to their neighbors. However, it is unclear how those properties relate to the learning ability, which we will explore in the future work. 

Next, we demonstrate the advantage of  biaffine between ego and local-field, compared with "ego to ego" biaffine and "local-field to local-field" biaffine. To implement the "ego to ego" mode in our BAGCN framework, we replace the local-field representation module with FC layer. Similarly, for the "local to local" biaffine manner, we use two one-layer GCN modules for mutual attention. From the results, in line with our expectations, we believe that only using a single self-information or local information interactive calculation, can not well obtain long-distance node dependencies. Table ~\ref{tb:Loc-mul} and table~\ref{tb:Loc-add} show the results of the two variants of biaffine mapping ("Ego to Ego" and "Loc. to Loc.") on four datasets with different feature fusion operations, i.e., addition and multiplication, respectively. We can see that biaffine mapping between ego to local-field brings much more gains to the performance of the model. The reason is that "ego to ego" attention discards the structural information about nodes, while "local-field to local-field" excessively depends on the neighbors, which may suffer from low level of local homophily. In contrast, the biaffine between ego to local field takes both the topological information and original node feature into account, resulting in better performance.

\begin{table}
\small
	\centering
	\caption{Ego and local field of BAGCN-mul on four graphs. Results are averaged over 20 repeated runs.}
	\begin{tabular}{ccccc}
		\toprule
		\textbf{Dataset} & \textbf{Cora} & \textbf{Citeseer} & \textbf{Cora-ML} & \textbf{Photo}\\
		\midrule
		Ego to Ego & 81.7 & 71.0 & 79.0 & 90.1\\
		Loc. to Loc. & 82.1 & 72.11 & 82.3 & 66.1\\
		\midrule
		BAGCN-mul & \textbf{83.7} &\textbf{72.6} &\textbf{82.7} & \textbf{90.7}\\
		\bottomrule 
	\end{tabular}
	\label{tb:Loc-mul}
\end{table}

\begin{table}
\small
	\centering
	\caption{Ego and local field of BAGCN-add on four graphs. Results are averaged over 20 repeated runs.}
	\begin{tabular}{ccccc}
		\toprule
		\textbf{Dataset} & \textbf{Cora} & \textbf{Citeseer} & \textbf{Cora-ML} & \textbf{Photo}\\
		\midrule
		Ego to Ego & 81.3 & 72.0 & 81.5 & 90.24\\
		Loc. to Loc. & 82.7 & 71.7 & 82.8 & 90.04\\
		\midrule
		BAGCN-add & \textbf{83.3} &\textbf{73.0} &\textbf{83.9}& \textbf{92.3}\\
		\bottomrule
	\end{tabular}
	\label{tb:Loc-add}
\end{table}
\subsection{Effects of training set size}
To evaluate the performance of the proposed model under a small number of training samples, we compare our results with feature based model MLP, vanilla GCN, attention based model GAT, simplified graph network model for oversmoothing SGC~\cite{2019sgc} and deep model APPNP. We use random training/validation/test splits for every training set size. To test whether BAGCN is capable of capturing the information of long-range neighbors, we conduct a suite of experiments by varying the number of training nodes from 50 to only 1. The results are averaged over 100 runs on Cora dataset. The results shown in Table~\ref{tb:train_size} suggest that, BAGCN outperforms the baseline models consistently under different settings of training set size. Notably, when there is only one labeled node for each class, BAGCN shows an overwhelming performance over GCN by a significant margin of 17.8\%. The considerable performance improvements are largely due to BAGCN's wide receptive field. In effect, it is essential for a model to exploit the structural information when the number of training nodes is very limited. Different from most of the existing GCNs that continuously pass  messages along the edges with multiple layers (e.g APPNP and SGC), BAGCN attempts to capture long-distant dependencies directly, which makes it possible to share feature/label information with the nodes of the same label that are distributed in discontiguous regions.

\begin{table*}[htbp]
\small
	\centering
	\caption{Results with different training set sizes on Cora in terms of classification accuracy (in percent).}
	\label{tb:train_size}
	\begin{tabular}{cccccccccc}
		\toprule
		\textbf{Training nodes per class} & \textbf{1} & \textbf{2} & \textbf{3} & \textbf{4} & \textbf{5} & \textbf{10} & \textbf{20} & \textbf{30} & \textbf{50} \\
		\midrule
		MLP &30.3 & 35.0 & 38.3 & 40.8 & 44.7 & 53.0 & 59.8 & 63.0 & 65.4  \\
		GCN &34.7 & 48.9 & 56.8 & 62.5 & 65.3 & 74.3 & 79.1 & 80.8 & 82.9  \\
		GAT &45.3 & 58.8 & 66.6 & 68.4 & 70.7 & 77.0 & 80.8 & 82.6 & 84.0  \\
		SGC &43.7 & 59.2 & 67.2 & 70.4 & 71.5 & 77.5 & 80.4 & 81.3 & 82.1  \\
		APPNP &44.7 & 58.7 & 66.3 & 71.2 & 74.1 & 79.0 & 81.9 & 83.2 & 84.3  \\
		\midrule
		BAGCN-mul & \textbf{52.5} & \textbf{60.5} & \textbf{70.9} & \textbf{72.2} & \textbf{74.8} & \textbf{79.7} & \textbf{82.7} & \textbf{84.0} & \textbf{84.5}  \\
		\bottomrule
	\end{tabular}
\end{table*}

\subsection{Ablation Study}

To evaluate the  effectiveness of biaffine attention, contrastive learning and sharpening, we conduct ablation studies for two variants of BAGCN (BAGCN-add and BAGCN-mul) on two large datasets, i.e., Cora-ML and Photo as Table~\ref{tb:ablation-mul} and Table~\ref{tb:ablation-add} show. In the first line of each dataset in two tables, we remove the contrastive learning loss from the total loss (i.e., $\lambda$ = 0); in the second line, we do not include biaffine attention but apply two-layer GCNs and MLP for contrastive learning, wherein sharpening trick is also used; in the third line, we implement BAGCN without sharpening (i.e., $\mathcal{T}$ = 1). The experiments show the notably benefits of the biaffine, contrastive learning and sharpening, but the biaffine module significantly improves over the vanilla contrastive learning of GCN and MLP, meaning that long-range dependencies between nodes are informative for node representation.

Moreover, to test whether the alignment with the average (represented by $\mathcal{L}_{con}$) is better than the alignment with each other, denoted by $CL2$ (i.e., $\left \| \boldsymbol{\hat{Y}_{gcn}} - \boldsymbol{\hat{Y}_{fc}}\right \|_{2}^{2}$), we perform BAGCN with $CL2$ regularization by replacing $CL$ loss with $CL2$. Compared with the last lines of two tables, the performance of the model deteriorates significantly, suggesting that the alignment with the average is better.

\begin{figure}
	\subfigure[]{
		\includegraphics[scale=0.23]{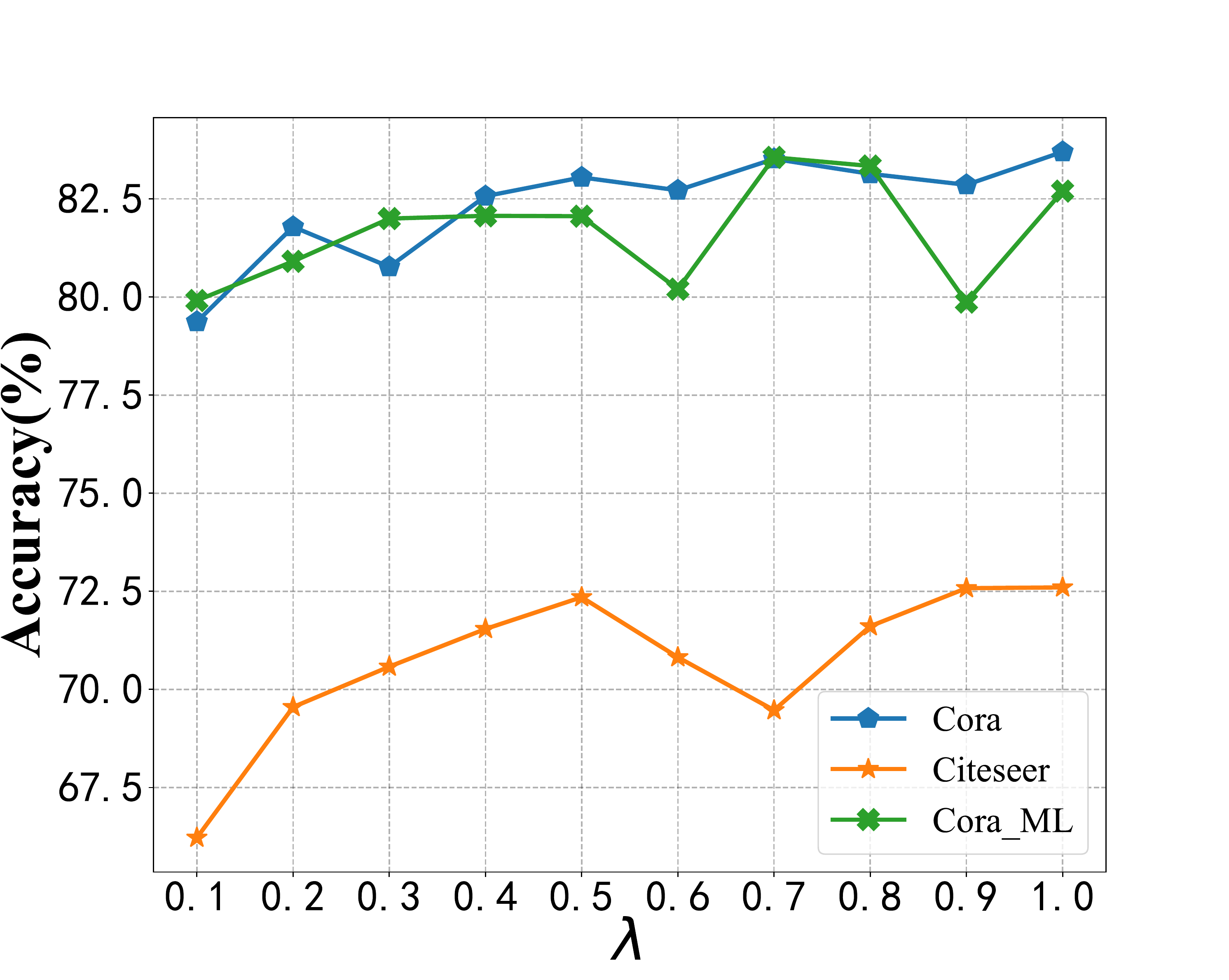} 
	}
	\subfigure[]{
		\includegraphics[scale=0.23]{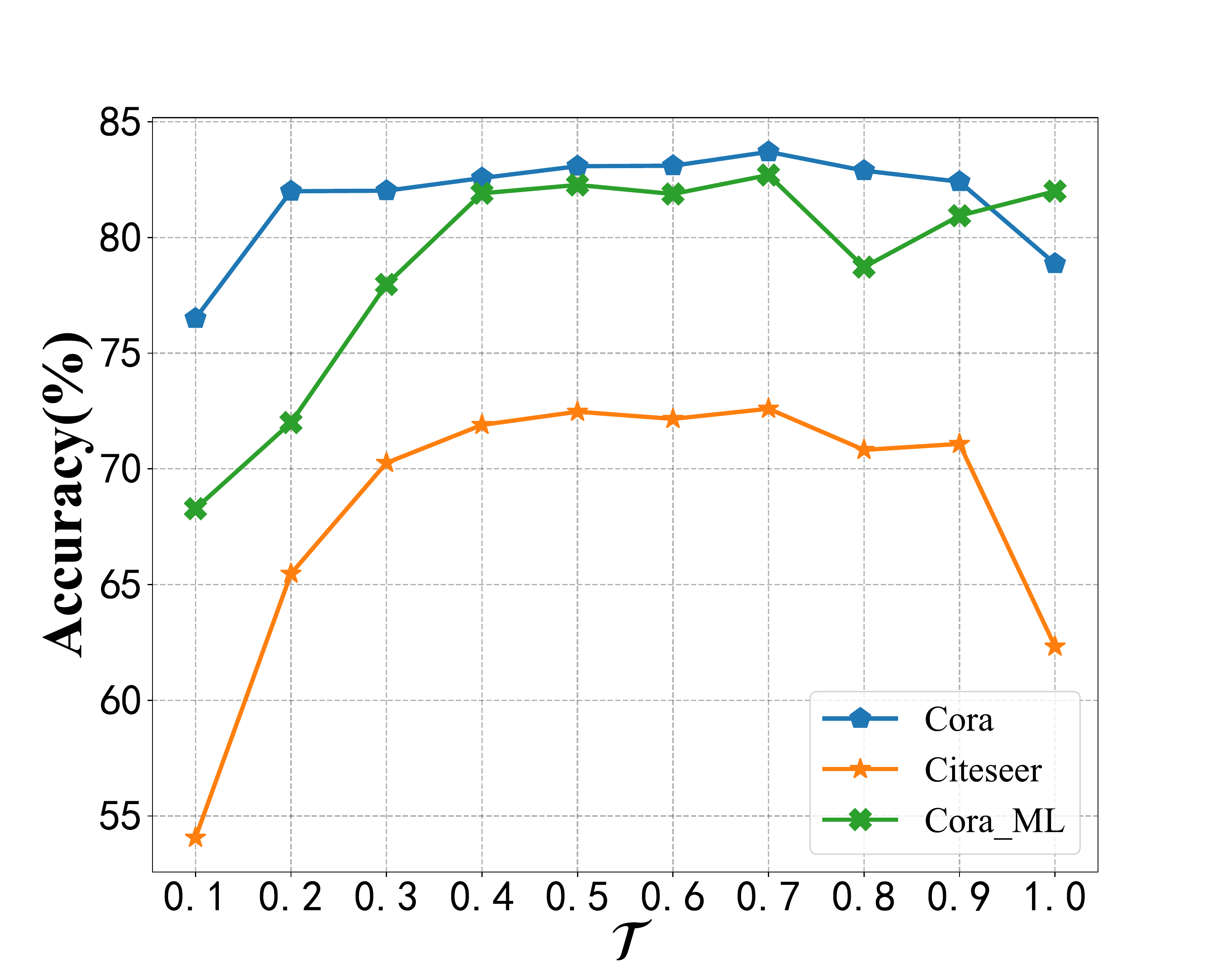}
	}
	\caption{(a) Regularization path of BAGCN-mul on Cora, Citeseer and Cora-ML. (b) The impacts of sharpening temperature on accuracy.}
	\label{fig:hyper-mul}
\end{figure}

\begin{figure}
	\subfigure[]{
		\includegraphics[scale=0.23]{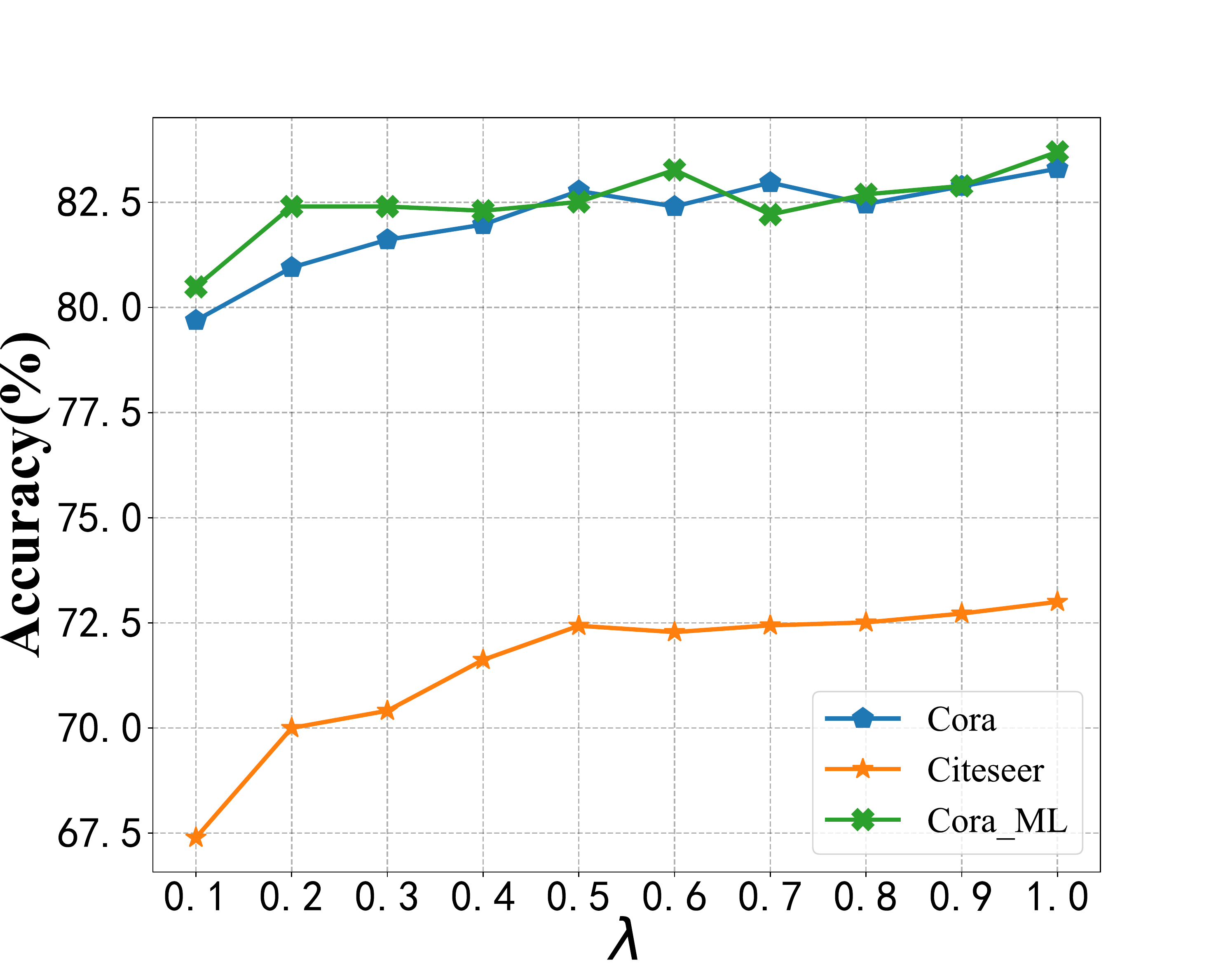} 
	}
	\subfigure[]{
		\includegraphics[scale=0.23]{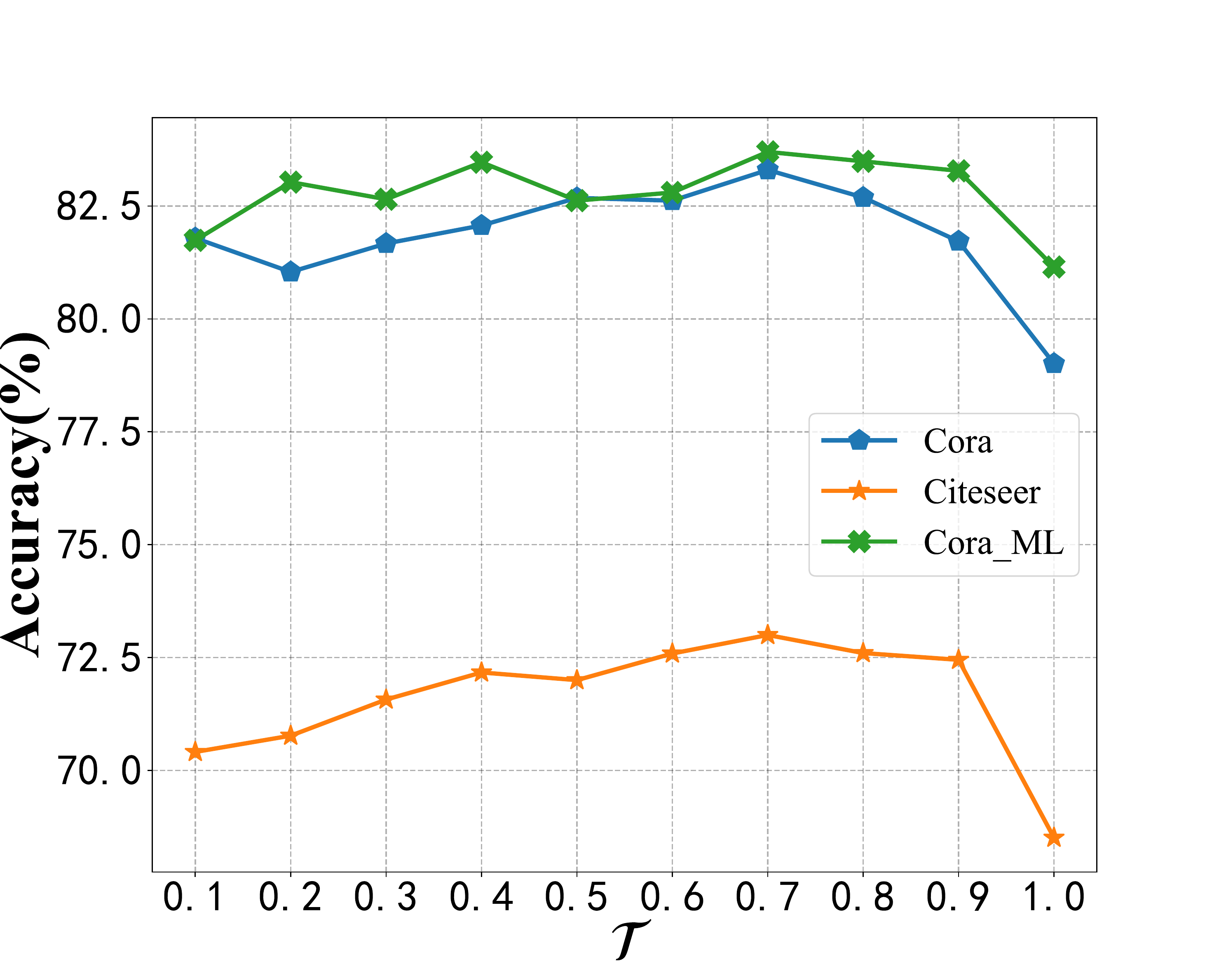}
	}
	\caption{(a) Regularization path of BAGCN-add on Cora, Citeseer and Cora-ML. (b) The impacts of sharpening temperature on accuracy.}
	\label{fig:hyper-add}
\end{figure}
\begin{table}
\small
	\centering
	\caption{Ablation study of BAGCN-mul on four graphs. Results are averaged over 20 repeated runs.}
	\begin{tabular}{ccccc}
		\toprule
		\textbf{Dataset} & \textbf{Cora}  & \textbf{Citeseer}& \textbf{Cora-ML} & \textbf{Photo} \\
		\midrule
		w/o CL & 79.3 & 64.4 & 80.0 & 90.6 \\
		w/o BA & 79.1 & 66.2 & 81.5 & 89.6 \\
		w/o S & 79.1 & 62.3 & 80.2 & 90.1 \\
		CL2 & 74.0 & 52.8 & 75.2 & 89.8 \\
		\midrule
		BAGCN-mul &\textbf{83.7} &\textbf{72.6} & \textbf{82.7} &\textbf{90.7} \\
		\bottomrule
	\end{tabular}
	\label{tb:ablation-mul}
\end{table}
\begin{table}
\small
	\centering
	\caption{Ablation study for BAGCN-add. Results are averaged over 20 repeated runs.}
	\begin{tabular}{ccccc}
		\toprule
		\textbf{Dataset}& \textbf{Cora}& \textbf{Citeseer}  & \textbf{Cora-ML} & \textbf{Photo} \\
		\midrule
		w/o CL & 79.0& 65.5& 82.3 & 91.0\\
		w/o BA & 73.7& 63.7& 81.5 & 90.1\\
		w/o S & 79.4& 67.6& 82.2 & 90.4\\
		CL2 & 74.5& 61.3& 74.9 & 89.7\\
		\midrule
		BAGCN-add & \textbf{83.3}& \textbf{73.0}& \textbf{83.9} & \textbf{92.3}\\
		\bottomrule
	\end{tabular}
	\label{tb:ablation-add}
\end{table}
\subsection{Hyper-parameters}

There are two hyper-parameters involved in the training phase, namely, the regularization factor $\lambda$ to weigh two-views' alignment and the temperature $\tau$ to tune the sharpness of the logits' distribution. Figure~\ref{fig:hyper-mul} shows the regularization paths on three benchmarks, indicating that equally treating the two components of the total loss will bring a better gain for the model. As to the temperature $\tau$, it is shown in Figure~\ref{fig:hyper-add} that larger temperature tends to perform passivation on the logits' distribution, i.e., all logits evenly distributed across the classes. But too sharpened distribution  corresponding to small $\tau$ is also not expected, as it suggests a one-hot prediction results. From the experimental results on three datasets, we observe that the model generally achieves promising performance when $\tau$ is around 0.7. Thus we set this parameter to 0.7 in our experiments.

\section{Conclusions}

In this work, we have addressed the limits of shallow GCNs in expressiveness and the shortcomings when resorting to deep architectures. To tackle this problem, we have proposed to build shortcuts between distant nodes by leveraging biaffine mapping technique. Based on the representations learned with biaffine transformation, we have devised a multi-view contrasting learning scheme for semi-supervised node classification, which allows the local field representation of the nodes to be adjusted by the original node features indirectly. We have conducted a suit of experiments to evaluate the performance of the proposed method BAGCN. The results demonstrate that it achieves the state-of-the-art performance on almost all the nine benchmark datasets of different scales. Moreover, BAGCN shows improved robustness compared to vanilla GCN, and consistently outperforms over four strong GCN models for different sizes of training. However, as biaffine mapping implements the global attention, the parameters to be learned  are more than in local attention. Our future work includes designing sampling-based bi-affine techniques to improve the efficiency of the model.

\section*{Acknowledgments}
 {
	This work is partially supported by National Natural Science Foundation of China (No.62776099) and SWPU Innovation Base No.642.
}

\bibliographystyle{IEEEtran}
\bibliography{IEEEtran}
\end{document}